\documentclass[lettersize,journal]{IEEEtran}
\usepackage{amsmath,amsfonts}
\usepackage{algorithmic}
\usepackage{algorithm}
\usepackage{array}
\usepackage[caption=false,font=normalsize,labelfont=sf,textfont=sf]{subfig}
\usepackage{textcomp}
\usepackage{stfloats}
\usepackage{url}
\usepackage{verbatim}
\usepackage{graphicx}
\usepackage{cite}
\usepackage{lscape} 
\usepackage{comment}
\usepackage[table]{xcolor}
\usepackage{booktabs}
\usepackage{pifont}
\usepackage{comment}

\begin{document}

\title{Vision-Based Mistake Analysis in Procedural Activities: A Review of Advances and Challenges}

\author{Konstantinos~Bacharidis and Antonis~A.~Argyros%
\thanks{Both authors are with the Institute of Computer Science, Foundation for Research and Technology – Hellas, Greece and the Computer Science Department, University of Crete, Greece (email: \{kbach, argyros\}@ics.forth.gr).}%
}


\markboth{Vision-Based Mistake Analysis in Procedural Activities: A Review of Advances and Challenges}
{Bacharidis and Argyros}


\maketitle

\begin{abstract}
Mistake analysis in procedural activities is a critical area of research with applications spanning industrial automation, physical rehabilitation, education and human-robot collaboration. This paper reviews vision-based methods for detecting and predicting mistakes in structured tasks, focusing on procedural and executional errors. By leveraging advancements in computer vision, including action recognition, anticipation and activity understanding, vision-based systems can identify deviations in task execution, such as incorrect sequencing, use of improper techniques, or timing errors. We explore the challenges posed by intra-class variability, viewpoint differences and compositional activity structures, which complicate mistake detection. Additionally, we provide a comprehensive overview of existing datasets, evaluation metrics and state-of-the-art methods, categorizing approaches based on their use of procedural structure, supervision levels and learning strategies. Open challenges, such as distinguishing permissible variations from true mistakes and modeling error propagation are discussed alongside future directions, including neuro-symbolic reasoning and counterfactual state modeling. This work aims to establish a unified perspective on vision-based mistake analysis in procedural activities, highlighting its potential to enhance safety, efficiency and task performance across diverse domains.
\end{abstract}


\section{Introduction}
\label{sec:introduction}

Making mistakes is intrinsic to human nature. While they can act as critical drivers for learning and skill refinement~\cite{reason1990human}, they may also lead to costly or dangerous consequences in high-stakes environments. The ability to detect, anticipate, and respond to mistakes, ideally before they escalate into failures, is essential in contexts where safety, efficiency, or performance are paramount. In such scenarios, real-time mistake analysis can support timely interventions and corrective actions, ultimately improving both human and system-level outcomes.

\begin{figure}[t!]
    \centering
    \includegraphics[width=0.95\linewidth]{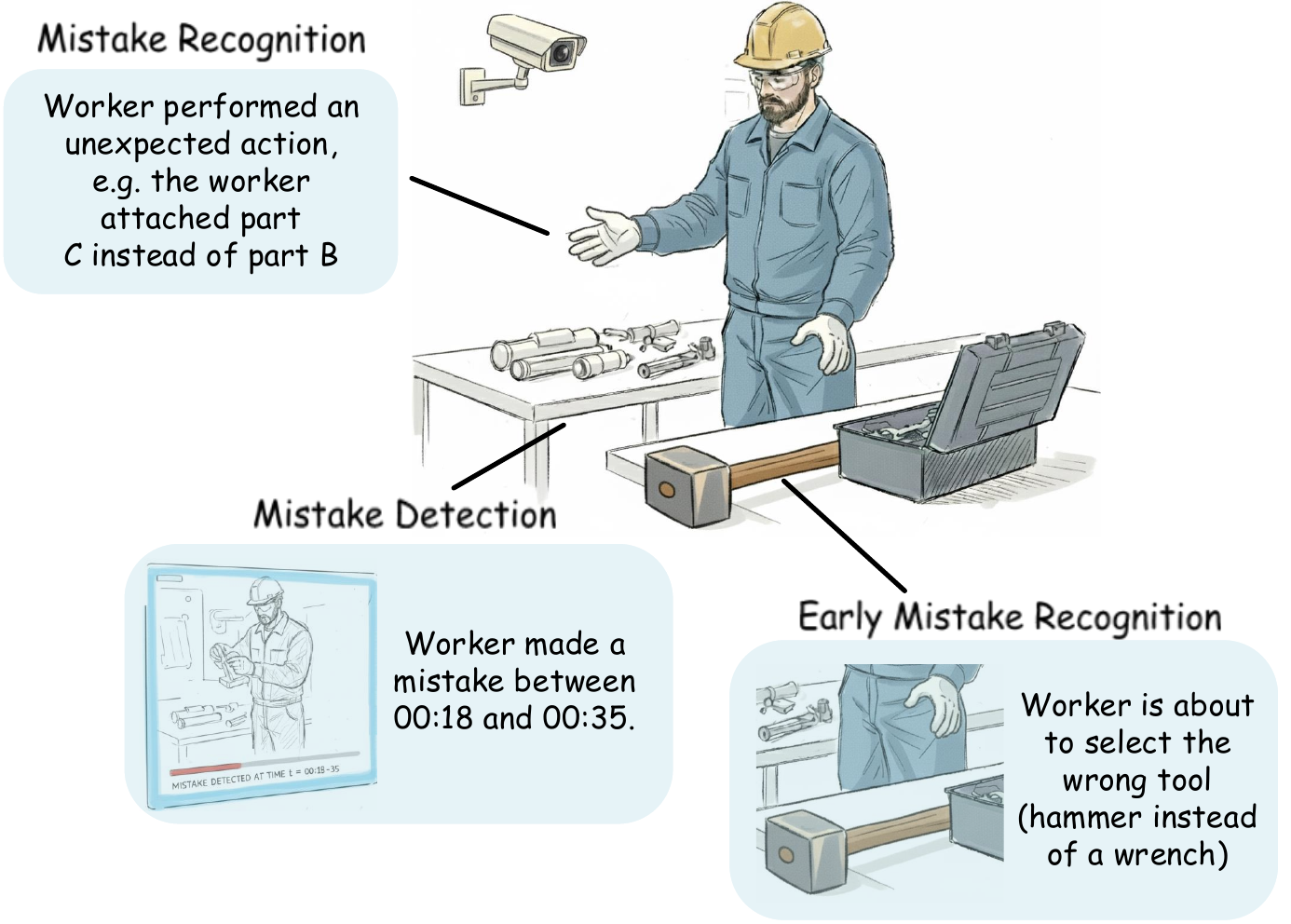}
    \caption{\text{Illustration of mistake analysis in an industrial assembly task:} (1) recognize an unexpected or omitted action given an activity protocol (\textit{mistake recognition}), e.g.,the worker ``attached part C'' instead of ``grabbing part B'', (2) temporally detect an error within the execution of an action (\textit{mistake detection}) and (3)~predict a mistake before it fully occurs (\textit{early mistake recognition}), e.g., warn the worker about selecting the wrong tool. \textcolor{black}{The conceptual worker sketch was generated with \cite{google_gemini}}.}
    \label{fig:concept}
\end{figure}

Mistake analysis can take on multiple, complementary forms. At times, the challenge lies in recognizing that a mistake has occurred (\textit{mistake recognition}); in others, it is about identifying precisely when it happens during the task (\textit{mistake detection}); and in critical situations, the key lies in anticipating mistakes early enough to prevent them altogether (\textit{early mistake recognition}). These perspectives naturally arise from the way humans reason about everyday errors and offer an intuitive framework for understanding how intelligent systems might tackle similar challenges.

The impact of robust mistake analysis methods spans multiple domains. In industrial automation, they can enhance productivity by minimizing human error, ensuring proper task execution, and reducing the risk of injuries (see Figure~\ref{fig:concept}). In human-robot collaboration, anticipating potential mistakes facilitates smoother cooperation, reduces failure rates, and improves task handover efficiency. Similarly, in rehabilitation, education, or assisted living, such systems can guide users through complex tasks, offering feedback and corrections that enhance skill acquisition or daily independence. In embodied robotics, mistake analysis plays a dual role: supporting human–robot interaction~\cite{nikolaidis2017human} and enabling robots to recognize and recover from abnormal or unexpected situations~\cite{garrabe2024enhancing}, a critical step toward robust autonomy in real-world environments

\begin{figure*}[t!]
    \centering
    \includegraphics[width=0.94\textwidth]{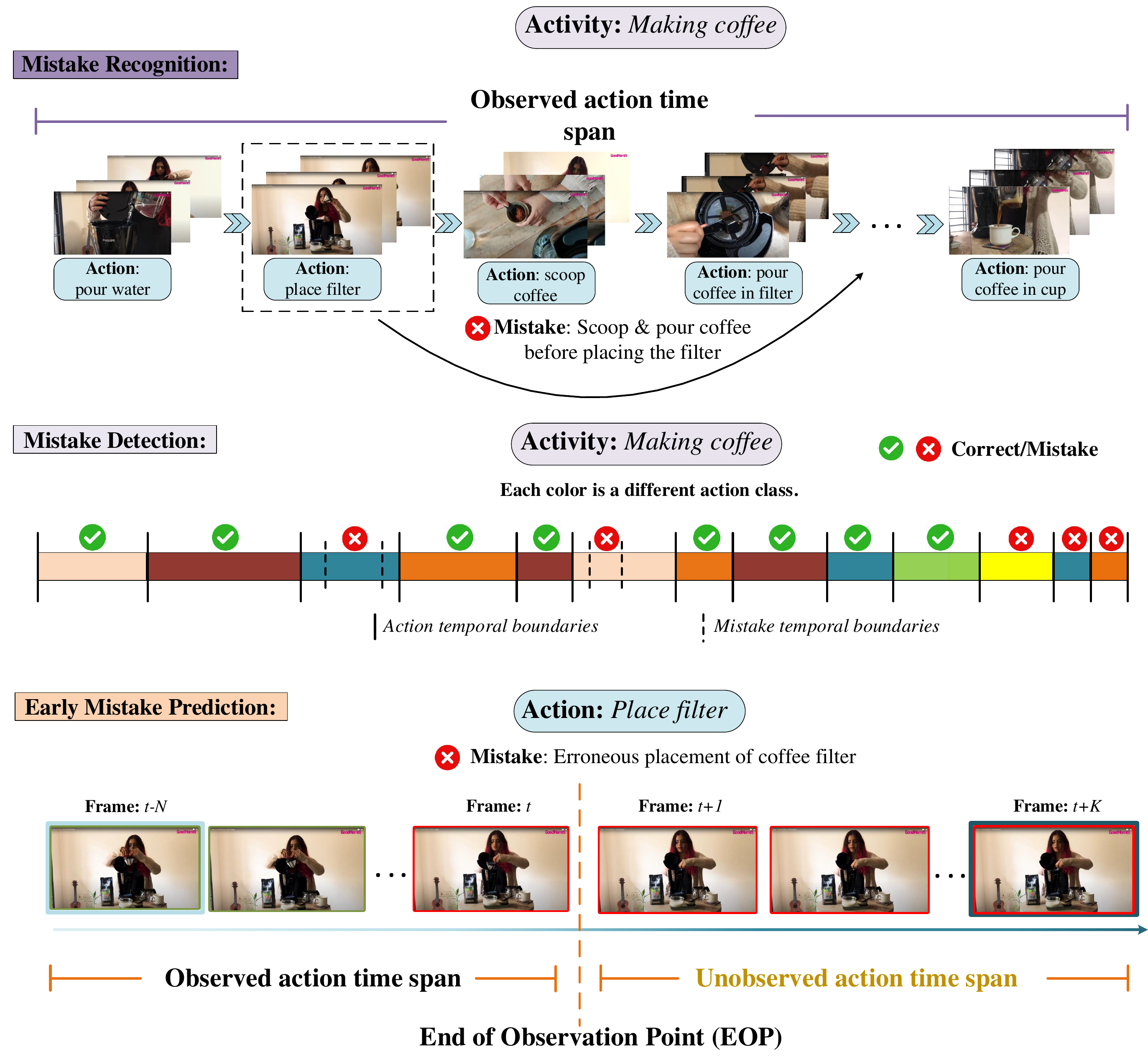}
    \caption{{Illustration of the differences between mistake recognition, early mistake recognition and mistake detection using an example from a filter coffee preparation video.} In mistake recognition, the error (e.g., omitting coffee grounds) is identified after the action is completed. In early mistake recognition, the model anticipates the mistake before it fully unfolds (e.g., detect intent to add coffee before coffee filter is added) in the on-going action. In mistake detection the model temporally localizes the specific segment where the mistake occurs. The temporal boundaries of the action and mistake segments can overlap.}
    \label{fig:tasks_overview}
\end{figure*}

A unifying aspect across these diverse scenarios and domains is that they all involve procedural activities, i.e. structured tasks composed of ordered, goal-driven action sequences. Whether in manufacturing, cooking, or caregiving, such procedures typically follow established protocols and depend on the correct execution of each step. This structure not only makes deviations easier to define and detect, but also means that even small mistakes, such as step skipping, executing actions in the wrong order, or using an incorrect tool, can significantly affect outcomes. Consequently, procedural activities provide a natural and impactful setting for mistake analysis: their regularity enables modeling of expected behaviors, while their error sensitivity underscores the value of timely detection and intervention. Understanding where and how procedures go off track supports more effective training, safer automation, and helps users maintain task success in real-world environments.

Computer vision has emerged as a powerful modality for mistake detection in procedural tasks. Unlike other sensor-based methods, vision offers a non-invasive, cost-effective, and scalable alternative leveraging rich visual data. Advances in activity recognition, action detection, and multimodal understanding~\cite{stergiou2025time} enable detecting subtle deviations such as incorrect motion trajectories, missing steps, or tool misuse. For instance, in cooking, a vision system might infer an error when a user prematurely switches tools or leaves a pan unattended.

Despite recent progress, mistake analysis in procedural activities remains highly challenging~\cite{aggarwal2011human,herath2017going}. A major difficulty lies in the variability of task execution: even well-defined procedures often allow multiple valid action sequences rather than a single canonical order. This complicates the distinction between genuine errors and acceptable alternatives. Further ambiguity arises at the action level, where steps vary in motion dynamics, timing, or appearance, making the boundary between atypical yet correct and truly erroneous behavior non-trivial. Additional challenges stem from viewpoint changes (e.g., egocentric vs.\ exocentric), object multifunctionality, and activity compositionality, which demand fine-grained spatiotemporal reasoning. Crucially, mistake detection requires judging whether an action is appropriate in context, making it a higher-order reasoning task beyond static recognition.

The increasing significance of this problem has driven recent progress in vision-based methods and benchmark datasets for mistake detection and anticipation. Yet, a unified survey remains lacking. To our knowledge, this is the first systematic review of video-based mistake analysis in procedural activities. We summarize existing datasets, categorize mistake types, outline evaluation protocols, and survey state-of-the-art recognition and prediction methods, offering a unified perspective and identifying key challenges and open questions.

\section{Mistake Analysis: Problem Statement}
\label{DefMis}
As mentioned in Section~\ref{sec:introduction}, the problem of video-based mistake analysis in procedural activities can be naturally decomposed into three interconnected sub-tasks, depicted in Figure~\ref{fig:tasks_overview}:  
(a)~\textbf{mistake recognition},  
(b) \textbf{mistake detection} and  
(c) \textbf{early mistake recognition}.  
Among these, recognition constitutes the foundational task, since it provides the basis upon which detection and early prediction are built. We therefore begin with a formal definition of mistake recognition and then extend this framework to detection and early prediction.
\textcolor{black}{In the definitions that follow, these tasks are treated as downstream problems that rely on action and activity understanding. Specifically, mistake analysis requires knowledge of both the performed actions and the associated high-level activities, as errors are defined relative to the expected execution of these actions within their activity context.}

\vspace{0.1cm}

\noindent{\bf Mistake recognition:} Given an untrimmed video of length $T$ frames, 
$\mathbf{x}_{1:T}=\langle x_1,\dots,x_T\rangle$
capturing the execution of an activity, 
the goal is to identify deviations from expected behavior, termed \textit{mistakes}. 
This task can be decomposed into two stages: 
a) \textit{action detection} and \textit{activity recognition}: divide the continuous video into temporally localized clips, each assigned an action and an activity label;  
b) \textit{mistake recognition}: reason about whether each temporal segment constitutes a deviation from the execution protocol of the assigned action and activity.

In more detail, in the first step of the process,  given an untrimmed video, an \textit{action detection} method produces a sequence of clips 
\( V = \{v_1, v_2, \dots, v_N\} \), 
where each clip \( v_n \) is associated with an action label \( a_n \in \mathcal{A} \), with \( \mathcal{A} \) denoting the set of possible actions. This is followed by an \textit{activity recognition} step which associates the sequence of detected actions with an activity label \( y_n \in \mathcal{Y} \), 
with \( \mathcal{Y} \) denoting the set of high-level activities. The final output of this stage is the annotated sequence 
\[
S = \{(v_n, a_n, y_n)\}_{n=1}^N,
\] 
which serves as input to the \textit{mistake recognition} stage.

Given the sequence \( S \) of action- and activity-labeled clips, 
the objective of the second stage (\textit{mistake recognition}) is to analyze \( S \) to identify specific segments that deviate from the expected behavior protocol \( \mathcal{B} \) and, when possible, characterize the type of deviation. These deviations, referred to as \textit{mistakes} or \textit{errors}, can be broadly categorized into:

\begin{itemize}
    \item \textbf{Procedural errors:} Actions performed in incorrect order, omitted, repeated, or performed at inappropriate times.
    \item \textbf{Execution errors:} Actions executed with incorrect motion, posture, or object manipulation relative to the expected pattern.
\end{itemize}

\noindent Specifically, the task is to learn a mapping, $f (\cdot)$, so that
\[
f: S \rightarrow M.
\]
\( M = \{m_i\}_{i=1}^N \) is a binary mistake label sequence with \( m_n \in \{0, 1\} \), where \( m_n = 1 \) denotes that clip \( v_n \) contains a mistake.

A more informative variant involves not only detecting whether a mistake occurs, but also classifying its type and subtype. Let \( C = \{c_1, c_2, \dots, c_N\} \) with \( c_n \in \mathcal{C} \), where \( \mathcal{C} \) is the set of mistake types (e.g., procedural or execution) and let \( E = \{e_1, e_2, \dots, e_N\} \) with \( e_n \in \mathcal{E}_{c_n} \), where \( \mathcal{E}_{c_n} \) denotes the set of mistake variants associated with type \( c_n \). The task then becomes a multi-level classification problem:
\[
f: S \rightarrow (M, C, E).
\]
Such rich output enables explainable and actionable feedback, facilitating effective interventions and correction strategies.

\vspace{0.1cm}
\noindent \textbf{Mistake detection:}  
Building on recognition, mistake detection generalizes the problem, by requiring both the temporal localization and classification of mistaken action segments to be inferred jointly. Specifically, given the sequence of temporally localized clips with associated action and activity labels from the first stage
\[
S = \{(v_n, a_n, y_n)\}_{n=1}^N
\]
the aim of the mistake detection task is to learn a mapping
\[
f: S \mapsto L, \quad 
L = \{([t^{(j)}_{\text{start}}, t^{(j)}_{\text{end}}], c_j)\}_{j=1}^m,
\]
where each segment $l_j$ denotes the temporal boundaries of a detected mistake and its associated type $c_j$. Mistake detection thus generalizes recognition by discovering error segments from the jointly inferred action/activity sequence rather than assuming pre-segmented clips, aligning the task with real-world procedural analysis pipelines.

\vspace{0.1cm}
\noindent \textbf{Early Mistake recognition:}  
The early recognition task extends recognition into the anticipatory setting, where the goal is to forecast mistakes before the completion of the corresponding procedural step.  
In more detail, let $v$ denote a video segment corresponding to a step of duration $T$ and define a partially observed prefix $v^{\text{partial}} = v_{1:t}$ with $t < T$. The model must learn a predictive mapping
\[
f: v^{\text{partial}} \rightarrow (M, C, E),
\]
where the output anticipates whether the full step $v$ will eventually contain a mistake and if so, its type and subtype.  
This predictive formulation introduces inherent uncertainty due to incomplete observations, 
yet it enables proactive intervention strategies that can prevent errors before they fully materialize.

\noindent A graphical illustration and disambiguation of the three subproblems of mistake analysis is provided in Figure~\ref{fig:tasks_overview}.

\begin{figure*}[t!]
    \centering
    \includegraphics[width=0.99\textwidth]{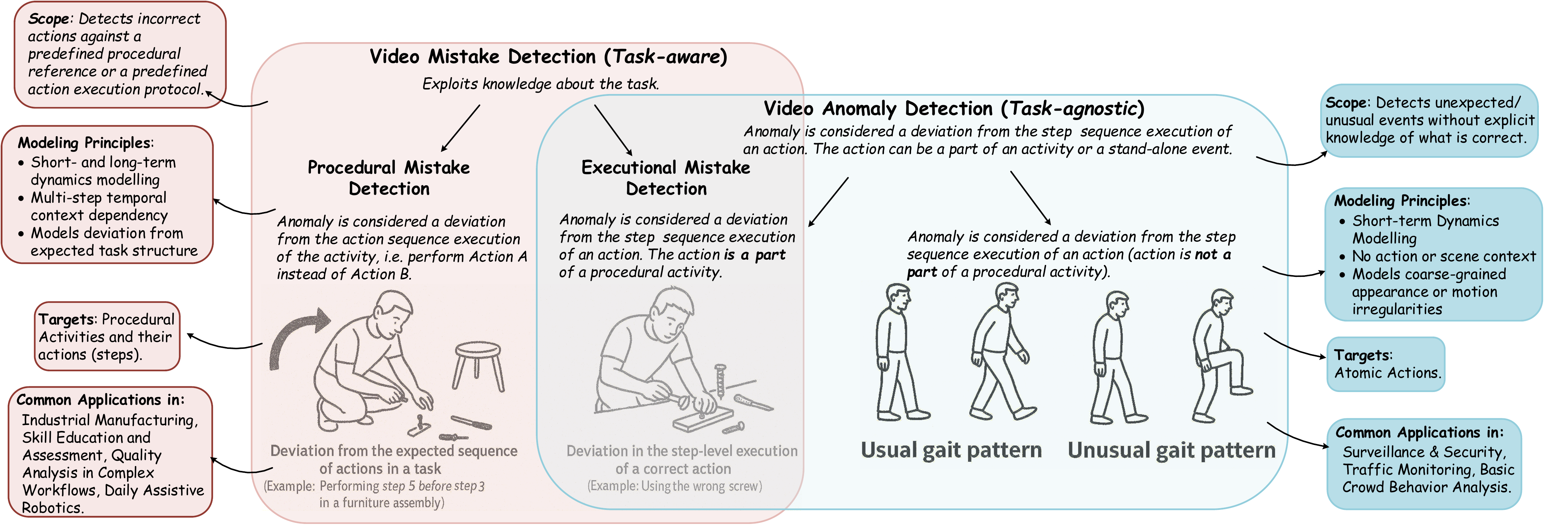}
    \caption{Illustration of the association between video anomaly and mistake detection in procedural activities.Human sketches in the figure generated with \cite{openai_chatgpt}.}
    \label{fig:anomaly_mistake}
\end{figure*}
\subsection{Distinguishing Mistake Analysis in Procedural Activities from Conventional Video Anomaly Detection}
\label{Mis_VAD}

Within the broader scope of mistake analysis, which encompasses mistake recognition, mistake detection, and early mistake prediction, we concentrate here on the task of mistake detection, as it can be viewed as a more generalized form of recognition, capturing not only whether a mistake occurs but also when and where it occurs within the activity. This temporal and spatial specificity makes detection a natural point of comparison to video anomaly detection, which similarly aims to localize unexpected events in continuous video streams.

In principle, mistake detection in procedural activities is related to video anomaly detection~\cite{V_Anomaly1,V_Anomaly2}. However, the two tasks differ fundamentally in the contextual grounding of what constitutes an error, with mistake detection requiring task-specific procedural knowledge to distinguish between acceptable variations and true deviations (Figure~\ref{fig:anomaly_mistake}).

Video anomaly detection typically focuses on identifying irregularities, i.e., deviations from normal visual or motion patterns, without necessarily understanding task-specific goals. Such anomalies exhibit distinct visual or semantic properties, making them recognizable as unexpected events. For instance, in the ShanghaiTech Campus dataset~\cite{ShanghaiTech}, a widely used benchmark, anomalies include unusual behaviors such as fighting, chasing, or other atypical activities, detected through their visual distinctiveness from typical campus scenes. Procedural mistake detection differs markedly from conventional anomaly detection. Unlike generic anomalies, mistakes in procedural activities are defined relative to a structured sequence of actions required to complete a task. These include missing, adding, modifying, or incorrectly executing an action step, all of which must be interpreted within the overarching goal. For example, an engineer assembling a circuit board who omits a resistor must disassemble parts to correct the mistake. Such errors stem from violating the prescribed action order and cannot be captured by appearance-based irregularities alone but require reasoning over long temporal dependencies. In contrast, video anomaly detection primarily flags unusual events based on appearance or motion, whereas mistake detection is inherently structured and goal-dependent.

A common ground between the two notions is executional mistakes in procedural activities, where the correct action is performed in a visually plausible way but with incorrect technique or suboptimal quality. This overlaps with conventional video anomaly detection since (a) both examine deviations at the atomic action level, (b) errors may still manifest as local visual irregularities (e.g., shaky hand movements, misaligned components, hesitant execution), and (c) both rely on modeling normal behavior to detect deviations that may not alter the overall sequence but affect execution quality or safety.

As a final remark, most anomaly detection methods operate on static or short-term temporal observations, whereas mistake detection in procedural activities demands long-range temporal reasoning to assess whether an action sequence aligns with a predefined procedure. This challenge is further amplified in egocentric video~\cite{EgoOops, CaptainCook4D, Assembly101}, where shifting viewpoints, occlusions and object-scale variations hinder the application of conventional anomaly detection techniques.

\section{A Taxonomy of Mistakes}

A fundamental step in mistake analysis for procedural activities is establishing an error-type taxonomy. Broadly, mistakes fall into two \textit{coarse-grained} categories: \textbf{procedural mistakes}, arising from deviations in the sequence relative to the activity protocol, and \textbf{executional mistakes}, occurring when an action is performed incorrectly in terms of motion quality, temporal fidelity, or object manipulation. Each category can be further decomposed into \textit{fine-grained} subcategories.

This organization naturally extends the binary problem of mistake recognition into a multi-class classification setting, where each detected mistake is assigned a coarse-grained label $c_i \in \mathcal{C}$ (procedural or executional) and a fine-grained subtype $e_i \in \mathcal{E}_{c_i}$. This layered formulation, referred to as \textit{error category recognition}, enables more detailed analyses of deviations, providing actionable insights into the causes and nature of mistakes in structured activities.  
\begin{figure*}[t!]
    \centering
    \includegraphics[width=0.99\linewidth]{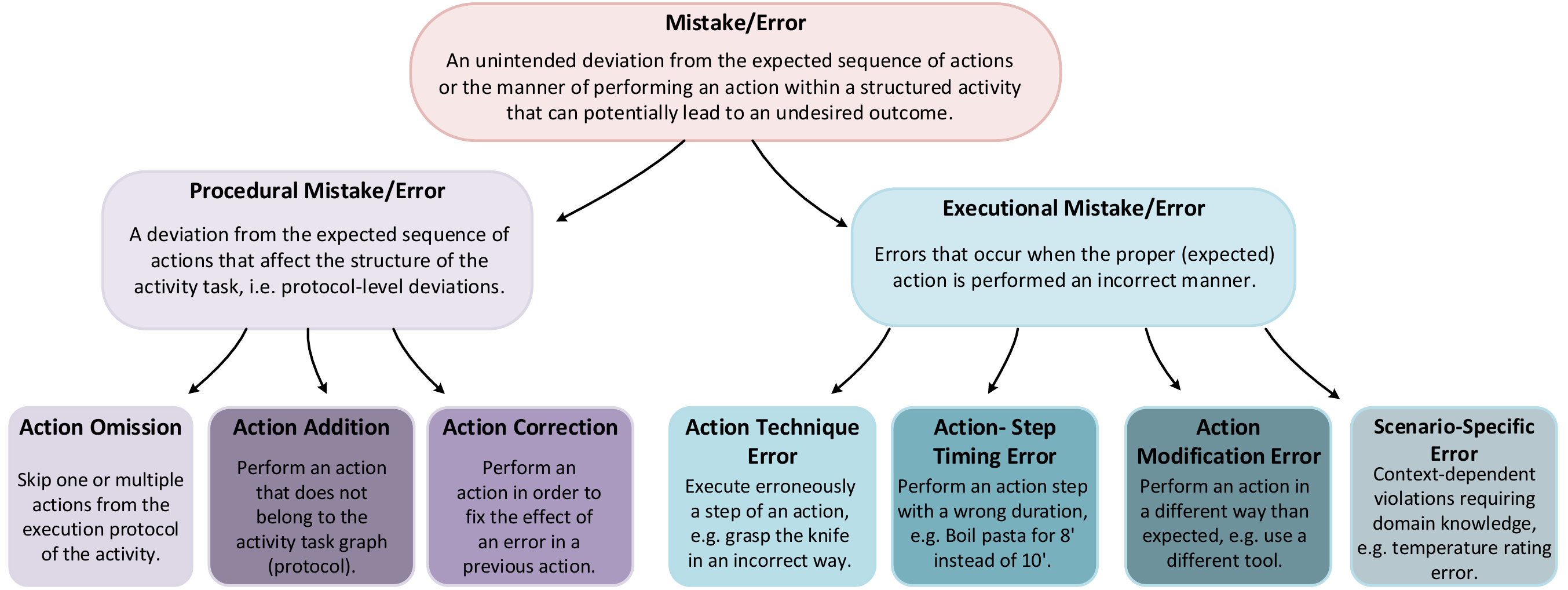}
    \caption{{Hierarchical taxonomy of error types in procedural activities}. The taxonomy distinguishes between procedural and executional errors, which arise from incorrect technique, timing, or action semantics. Scenario-specific errors are included to account for domain-dependent cases.}
    \label{fig:error_tax}
\end{figure*}
Due to the emerging nature of the problem, however, existing works adopt varied definitions of mistake/error types, particularly for executional mistakes. For instance, Peddi et al.~\cite{CaptainCook4D}, in the CaptainCook4D dataset, define the categories \{\textit{Order Error, Omission Error, Technique Error, Timing Error, Temperature Error}\}, with the first two being procedural and the remainder executional. In contrast, Lee et al.~\cite{EgoPED} propose a different but partially overlapping scheme, distinguishing \textit{Step Omission}, \textit{Step Addition} and \textit{Step Correction} as procedural, while identifying \textit{Step Modification} and \textit{Step Slip} as executional. Although both taxonomies include omissions and technique-based deviations, their definitions differ in granularity and interpretation.  A general overview of the taxonomy adopted in this survey is illustrated in Figure~\ref{fig:error_tax}. Following the distinctions introduced in prior work~\cite{ding2023mistakecountsassembly}, \textit{procedural mistakes} can be further divided into the following cases:
\begin{itemize}
    \item \textbf{Omitted actions:} an expected step is skipped.  
    \item \textbf{Unnecessary actions:} an extra known step outside the execution protocol is performed.  
    \item \textbf{Corrective actions:} an unexpected step is introduced to compensate for a prior mistake (only identifiable under an offline setting).  
\end{itemize}

In contrast, the categorization of \textit{executional mistakes} is inherently more nuanced and task-dependent, since their definition depends on the characteristics of the actions and constraints of the operational domain. Nevertheless, they can generally be grouped into four broad categories, each reflecting a different dimension of execution fidelity:
\begin{itemize}
   \item \textbf{Action technique errors} refer to discrepancies in how an action is performed. The procedural structure is preserved, actions occur in the correct order with the appropriate objects, but execution deviates from the expected technique, e.g., improper grip, incorrect alignment.
    \item \textbf{Temporal execution errors} capture deviations related to the \textit{timing or rhythm} of action execution. These include premature or delayed step initiation, violations of required step duration, or incorrect pacing that compromises task performance. For instance, removing an object from a heat source too soon may yield a functionally incorrect result even if all other steps are correctly executed.
   \item \textbf{Action modification errors} occur when an action is performed via an alternative approach deviating from the predefined protocol, such as using a different tool, skipping a sub-step, or combining multiple steps. Even if the intended outcome is achieved, the deviation introduces inconsistencies or risks relative to the standard procedure.
    \item \textbf{Scenario-specific errors} are context-dependent, tied to the semantics and requirements of particular actions within an activity. They occur when execution violates domain-specific constraints not captured by general categories like technique or timing. For instance, in surgery, placing a suture in the wrong anatomical region is a critical error, even if technique and timing are correct.
\end{itemize}

\section{Mistake Analysis Datasets}
\label{sec:Datasets}

\begin{figure*}[t]
    \centering
    \includegraphics[width=0.97\textwidth]{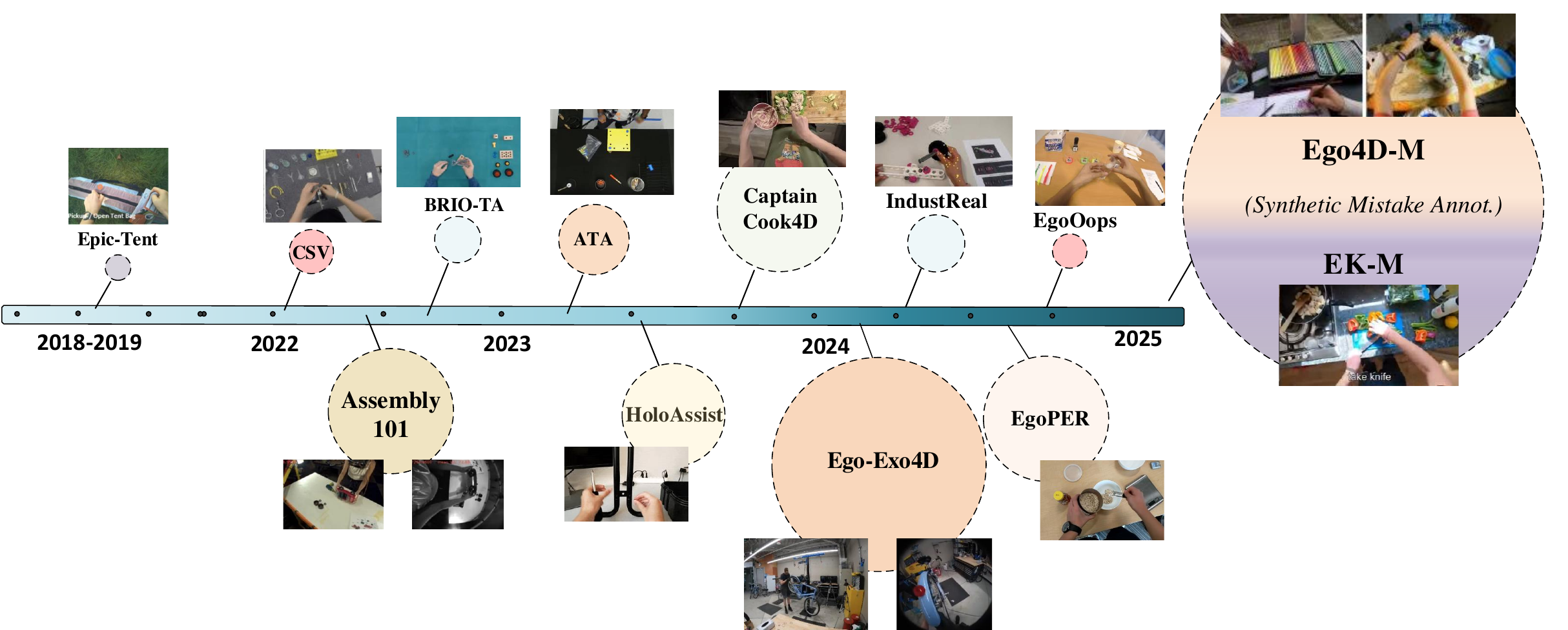}
        \caption{{Timeline of datasets supporting mistake analysis (2018–2025)}. Circle sizes refer to dataset scale. Datasets: Epic-Tent~\protect\cite{EPICTENT}, CSV~\protect\cite{CSV}, Assembly-101~\protect\cite{Assembly101}, BRIO-TA~\protect\cite{BRIOTA}, ATA~\protect\cite{ATA}, HoloAssist~\protect\cite{HoloAssist}, CaptainCook4D~\protect\cite{CaptainCook4D}, Ego-Exo4D~\protect\cite{grauman2022ego4d}, IndustReal~\protect\cite{IndustReal}, EgoPER~\protect\cite{EgoPED} and EgoOops~\protect\cite{EgoOops}. Ego4D-M and EK-M~\cite{li2025mistake} are synthetic extensions of Ego4D~\cite{grauman2022ego4d} and EK-100~\cite{damen2022rescaling}, generated to incorporate explicit mistake annotations.}
    \label{fig:data_time}
\end{figure*}

Only a handful of datasets introduced between 2018 and 2025 explicitly support mistake analysis in procedural activities, underscoring that this is an emerging research problem only recently addressed systematically. Figure~\ref{fig:data_time} presents a timeline and the relative size of these datasets.

\subsection{Datasets: overview \& key attributes}
\noindent Table~\ref{tab:Datasets} provides an overview of the datasets. Only tasks relevant to procedural action and activity understanding are included; tasks a dataset may support but not directly related to mistake analysis are omitted. The reported specifications refer to the subsets of these datasets used for such tasks, rather than the original dataset specifications, which may include additional vision tasks. All datasets under consideration include annotations for actions, activities, and supported error types, as well as additional ones that are specific to each dataset. 
\vspace{0.1cm}

\noindent\textbf{Epic-Tent~\cite{EPICTENT}:} Was the first dataset to offer annotated task-level mistakes in a real-world procedural activity setting. It comprises over 5.4 hours of first-person (ego-centric) video recorded from 29 participants, each equipped with two head-mounted cameras, tasked with assembling a tent in an outdoor setting. Epic-Tent includes rich multi-modal annotations, such as frame-level action and error labels, 2D gaze positions, and self-reported uncertainty levels, enabling a range of research applications in egocentric action and activity recognition, human-object interaction, and mistake detection. It supports coarse-to-fine procedural understanding, with the overall tent assembly task decomposed into 12 sub-activities, each comprising of low-level steps from a total of 38 unique actions. 

For mistake analysis, Epic-Tent provides coarse, segment-level annotations indicating whether the overall task or its sub-components were completed correctly or incorrectly, without distinguishing between procedural (e.g., skipping a step, incorrect order) and executional (e.g., misplacing a pole, using incorrect force) errors, though both may occur. Unlike other datasets targeting mistake analysis, Epic-Tent did not define evaluation protocols for this task, as nearly every sample included at least one procedural error, preventing binary success/failure splits. To address these limitations and explicitly enable online mistake detection, Flaborea et al.~\cite{flaborea2024prego} introduced \textit{Epic-Tent-O}, proposing an uncertainty-based dataset partitioning strategy: videos from high-confidence participants were used for training, while those from less confident performers, more prone to errors, were reserved for testing.
\vspace{0.1cm}

\noindent\textbf{Chemical Sequence Verification (CSV)~\cite{CSV}:} CSV addresses the challenges of procedural activity understanding and mistake analysis in the context of chemical procedures. It encompasses 14 activities, each representing a chemical experiment conducted by 82 volunteers following predefined scripts. The dataset comprises a collection of 70 video recordings, partitioned into train-test subsets under a 80-20 split. The recorded sequences include deviations from the predefined execution protocols, incorporating step-level transformations such as additions, deletions and reordering of procedural steps. Annotations follow a weak supervision scheme, providing video-level activity and action labels without temporal boundaries for the actions. Notably, the dataset does not include explicit annotations specifying the procedural mistake types. The dataset supports multiple tasks, including activity, action and mistake recognition. Additionally, it introduces the task of early mistake recognition (early warning) to predict errors before they fully manifest, enabling proactive intervention.
\vspace{0.1cm}

\noindent\textbf{Assembly101~\cite{Assembly101}:} This is a large-scale benchmark designed for action and activity understanding in assembly tasks, with a focus on fine-grained procedural activities. It contains over 362 videos which amount for 6,000 video clips of individuals assembling 101 toy car models, captured from multiple cameras (exocentric and egocentric) and viewing angles in a constrained laboratory environment.  Assembly101 supports a wide range of tasks, including fine-grained action recognition, action localization, action anticipation, activity recognition and mistake or error detection. Additionally, the dataset is designed to enable evaluation in zero-shot settings, allowing models to generalize to previously unseen actions or activity compositions. Each video is annotated with frame-level action annotations under two granularity levels (\textit{1380 fine-} (single-task) and \textit{202 coarse}-grained (multiple fine-grained)) and mistake annotations, making it suitable for both step-level and task-level reasoning. The dataset provides multi-modal data, including synchronized RGB video from multiple viewpoints, depth information and 3D hand pose data.

For mistake analysis tasks, annotations exist for 1.01K of the 6K benchmark video clips, provided at a coarse action-segment level, focusing on procedural errors, such as missing, or out-of-order steps. While executional errors (e.g., incorrect orientation or grasp) may occur in the data, they are relatively rare and not annotated. The annotation process does not differentiate between procedural and executional errors and no formal taxonomy of the mistake types is provided to support fine-grained mistake recognition. Subsequent works by Ding et al.~\cite{ding2023mistakecountsassembly} and Flabora et al.~\cite{flaborea2024prego} extended Assembly101 to better support the mistake detection task with additional annotations and better restructuring. Specifically, Ding et al.~\cite{ding2023mistakecountsassembly} enriched Assembly101 with  mistake-type information and explicit part-to-part connection details to facilitate procedural and mistake reasoning, whereas Flabora et al.~\cite{flaborea2024prego} enabled the support of \textit{online mistake detection}, producing the variant \textit{Assembly101-O}. In this adaptation, all correctly executed sequences were moved to the training set, while sequences with mistakes were reserved for validation and testing. Procedure lengths were adjusted to better suit real-time detection requirements.
\vspace{0.1cm}

\noindent\textbf{BRIO Toy Assembly (BRIO-TA)~\cite{BRIOTA}:} The dataset comprises 75 video recordings capturing both normal and anomalous assembly processes of a toy car model. The dataset includes three distinct error types that may arise during the assembly process: (a) step ordering variation, (b) step omission and (c) abnormal duration. Each erroneous sequence exhibits only one of these three anomaly types.  
The dataset provides segment-wise temporal annotations of action occurrences along with video-level mistake annotations. It supports the tasks of action recognition and segmentation, as well as mistake recognition. RGB videos is the only supported modality. Performance is evaluated using standard action segmentation metrics, including Accuracy, Intersection over Union (IoU) and Edit Distance, alongside mistake recognition metrics such as Accuracy. However, the dataset does not explicitly define an evaluation protocol or metrics tailored to each specific error type. 
\vspace{0.1cm}

\noindent\textbf{Anomalous Toy Assembly (ATA)~\cite{ATA}:} The design of the ATA dataset was motivated by the challenge of identifying mistakes in procedural activities, where the absence of fine-grained temporal annotations poses significant difficulties. The dataset comprises of 141 untrimmed videos of 32 participants assembling 3 different toy models, recorded from 4 distinct viewpoints. It includes both correct and erroneous executions, where errors can be seen or unseen during training, making it suitable for evaluating generalization in mistake detection. ATA is annotated with weak supervision in the form of video transcripts describing action sequences, while errors are not explicitly labeled in the transcripts, simulating real-world conditions where mistake annotations are unavailable. The captured errors encompass execution mistakes (e.g., misplacement of parts) and procedural mistakes (e.g., step omissions or order deviations), making it a comprehensive benchmark for evaluating mistake detection methods for both supervised and zero-shot settings. The dataset provides multiple modalities, including RGB videos, ASR (automatic speech recognition) transcripts and gaze data, enabling multimodal learning. 
\vspace{0.1cm}

\noindent\textbf{HoloAssist~\cite{HoloAssist}:} The HoloAssist dataset was designed to advance interactive AI assistants in real-world settings. It captures 20 collaborative object-centric manipulation tasks between two individuals: a \textit{task performer}, who executes the task while wearing a mixed-reality headset that records seven synchronized data streams, and a \textit{task instructor}, who observes the performer’s first-person feed in real time and provides verbal guidance, including interventions in case of mistakes. The dataset includes 166 hours of interaction data from 350 instructor–performer pairs drawn from 222 participants. Manual annotations accompany the dataset, including text summaries, intervention types, mistake labels, and segment-level action annotations at two abstraction levels: (a)~coarse-grained, describing high-level task steps, and (b)~fine-grained, referring to atomic actions composing each step. There are 414 coarse-grained and 1,887 fine-grained action classes.

For mistake analysis, HoloAssist provides segment-level binary labels denoting correct or erroneous action execution. It contains both procedural and executional errors, though without distinguishing between types. Notably, it introduces intervention annotations, a novel direction for AI assistive scenarios, covering three intervention types: correcting errors, following up with more instructions, and confirming the previous action. These are temporally aligned with the conversation transcripts, enhancing explainability in mistake analysis.
\vspace{0.1cm}

\noindent\textbf{CaptainCook4D~\cite{CaptainCook4D}:} This is one of the largest and most well structured datasets targeting the understanding of mistakes/errors in complex procedural activities within the domain of cooking. It comprises 384 recordings totaling over 94.5 hours of real-world kitchen interactions, capturing both standard and intentionally erroneous executions of 24 recipes. The activities are structured from a pool of 352 coarse-grained actions which are, in turn, defined from a set of fine-grained actions (steps). A standout feature of CaptainCook4D is its multimodal design, which includes RGB video, depth data, 3D object annotations, gaze data and audio. Unlike previous datasets that either lacked error annotations or only considered a binary notion of correctness, CaptainCook4D offers a rich and fine-grained taxonomy of 7 mistake types that capture both procedural and executional deviations. These mistakes are annotated at both the coarse- and fine-grained levels, allowing models to reason not only about what went wrong but also when and how this occurred. 
\vspace{0.1cm}

\noindent \textbf{Ego-Exo4D~\cite{grauman2024ego}:} This dataset was designed to support research in ego-exo video learning and multi-modal perception, focusing on skilled human activities. It is among the largest publicly available datasets containing time-synchronized first- and third-person video data, collected from 740 participants across 123 scenes in 13 cities worldwide. Participants performed 43 skilled physical and procedural activities unscripted within natural settings. In addition to RGB video, the dataset provides multi-modal sensory data, including audio, IMU, eye gaze, RGB and grayscale SLAM recordings, and 3D environment point clouds. It also includes time-indexed video-language resources, such as participant narrations during task execution and spoken expert commentaries evaluating performance. The dataset supports tasks related to activity understanding, including modeling instructor-learner relationships, recognizing fine-grained key-steps and task structures, assessing skill proficiency, and recovering 3D body and hand movements from egocentric video. Apart from action(step) and activity annotations, it provides 2D/3D pose annotations and object annotations, with an average of 5.5 objects annotated with correspondences between the two views in each take. Mistake analysis is introduced as a downstream problem within the task of procedure understanding. For this purpose, only 6 activities are considered, covering 186 step annotations. In total 628 sequences are recorded. The focus is on detecting procedural mistakes by identifying missing steps and recognizing steps that were not intended for the given activity.  
\vspace{0.1cm}

\noindent\textbf{IndustReal~\cite{IndustReal}:} This dataset was created to support the task of procedure step recognition in the context of industrial settings, with a particular focus on handling execution errors. The dataset contains 84 egocentric (first-person) videos, from 27 participants, captured during the assembly of a toy car, where the primary goal is to recognize and understand individual steps of the assembly process and detect execution errors within these steps. IndustReal focuses on two activities (a) the activity of assembling a toy car and (b) the activity of maintenance. The first is the most challenging, with the task being divided into 5 sub-activities, each containing multiple actions (steps). The total set of distinct actions in the dataset is 75. The structured breakdown of the assembly activity enables the tracking of task progression by identifying which sub-activity (or sub-task) has been completed. The dataset is enriched with multimodal sensor data, including video recordings and depth information (from stereo or depth cameras), providing a richer understanding of spatial and temporal context.

For mistake analysis, apart from annotations for procedural deviations (e.g., action omissions), IndustReal also includes executional error cases. For this error type, the dataset provides frame-level annotations for errors occurring during assembly at the action level, such as incorrect object interaction. In total, it contains 38 errors, 14 of which are exclusive to the validation and test sets. It was the first dataset to relate mistake analysis and assembly state progression, using sub-task completion as a means for fine-grained procedural mistake detection.
\vspace{0.1cm}

\noindent\textbf{EgoPER~\cite{EgoPED}:} EgoPER is a recent dataset that introduces a multi-modal benchmark for the study of mistake/error detection in egocentric procedural tasks. The dataset features 5 diverse cooking activities {\textit{preparation of pinwheels, coffee, quesadilla, tea and oatmeal}} performed by 11 participants. It contains over 120 hours of video recordings captured with a head-mounted camera in unconstrained environments. Unlike prior datasets that focus solely on successful task completions and generalized mistake recognition, EgoPER explicitly incorporates a rich taxonomy of human errors, including step omission, addition, modification, slip, and correction. Although all mistakes/errors are scripted, the execution of those errors was left natural, meaning participants had flexibility in how they introduced the error, leading to realistic and diverse instantiations of each error type. Each video is annotated with segment-wise action labels, object bounding boxes, hand-object interactions, gaze tracking, and audio, offering a comprehensive view of both correct and erroneous executions.
\vspace*{0.1cm}

\noindent\textbf{EgoOops~\cite{EgoOops}:} The dataset consists of 40 video recordings from 5 diverse task domains, namely \textit{electrical circuit assembly, color mixture experiments, ionic reactions, toy block construction and cardboard crafts}, captured in authentic but constrained (in terms of environment), educational and workshop settings. A key feature of EgoOops, shared only by CaptainCook4D and EgoPER, is procedural text integration, with texts tightly aligned with video segments to facilitate step-level grounding. EgoOops offers segment-wise annotations, including video-text alignments, detailed mistake labels (covering order and execution mistakes), object label annotations (microQR) and natural language descriptions of why specific actions deviate from the intended procedure, allowing for mistake explainability. Procedural mistakes in EgoOops follow the common ordering error scenarios found in related datasets, such as skipped, swapped, or repeated steps. In addition, the dataset introduces a detailed taxonomy of 6 executional mistakes, which occur when participants fail to correctly follow instructions in terms of executing the steps of an action (e.g., grasp wrong object). Finally, participants were instructed to follow scripted task and error scenarios, ensuring controlled coverage of error types; however, unintentional errors also occurred naturally during execution and were also annotated.
\vspace{0.1cm}

\noindent\textbf{Ego4D-M \& EPIC-KITCHENS-M:} Addressing the scarcity of large-scale mistake data, Li et al.~\cite{li2025mistake} introduced EK-M and Ego4D-M, which were synthetically created by repurposing existing egocentric corpora via a data engine named \textit{MisEngine}.  Unlike datasets relying on staged or naturally occurring errors, these benchmarks are generated by systematically cross-matching instruction texts with video segments of disparate actions (e.g., pairing a ``pick up hammer'' instruction with a ``pick up bolt'' video) based on Semantic Role Labeling (SRL) mismatches. This automated misalignment process allows for the inheritance of rich annotations from the source datasets (Ego4D~\cite{grauman2022ego4d} and EPIC-KITCHENS(EK)~\cite{damen2022rescaling}), resulting in benchmarks that are two orders of magnitude larger than prior works. Specifically, they provide approximately 257K and 221K samples respectively, covering $12,283$ and $16,099$ action classe respectively, offering supervision for fine-grained semantic, temporal (frame-level PNR~\footnote{Point-of-No -Return: signifies the reference frame for the mistake.}), and spatial (bounding box) mistake attribution.

\subsection{Datasets: Comparison \& Discussion}
\label{subsec:Compare}
The recent surge in datasets targeting procedural activity understanding and mistake analysis has considerably enriched the field, each offering distinct perspectives in terms of domain coverage, annotation granularity, sensing modalities and supported tasks. Datasets such as Assembly101, EpicTent and CaptainCook4D focus on structured multi-step activities (e.g., assembly or cooking), while others like EgoOops and CSV capture the variability of everyday tasks. Industrial and instructional contexts are represented by datasets such as EgoPER, ATA, IndustReal and BRIO-TA, while Holo-Assist and Ego-Exo4D emphasize AR guidance and ego-exocentric coordination. However, no single dataset fully captures the multifaceted nature of real-world mistake understanding. A deeper look shows that while individual datasets address certain challenges effectively, crucial aspects of the problem space remain underexplored or only partially covered. In what follows, we discuss several such aspects.

\begin{table*}[h!]
\centering
\scriptsize 
\setlength{\tabcolsep}{2pt} 
\renewcommand{\arraystretch}{1.1} 
\resizebox{\textwidth}{!}{ 
\begin{tabular}{p{2cm} | c | p{2.1cm} | c | c | c | p{1cm} | p{0.8cm} | c | c | c | c | c | p{2cm}}
\hline
\textbf{Dataset} & \textbf{Year} & \textbf{Domain}/ \textbf{Environment} & \textbf{View} & \textbf{\#Activ} & \textbf{\#Actions} & \textbf{Tasks} & \textbf{Error Type} & \textbf{\#Objs} & \textbf{3DM} & \textbf{\#Seqs.} & \textbf{Dur.} & \textbf{\#Partps.} & \textbf{Modalities \& Annotations} \\
\toprule
\midrule
EpicTent \cite{EPICTENT} & 2018, 2023 & Assembly/Real & Ego & 1 & 38 & MR & PEs$^{**}$ & NA & \ding{55} & 24 & $>$5.4h & 29 & RGB, Gaze \\
\hline
CSV \cite{CSV} & 2022 & Chemistry/Lab & Ego & 14 & 106 & MR, SR, EMR & PEs & NA & \ding{55} & 70 & 11.1h & 82 & RGB \\
\hline
Assembly101 \cite{Assembly101} & 2022 & Toy Assembly/Lab & Ego, Exo & 101 & 202 & MR, ASD, SR & PEs, EEs$^*$ & 15 & \ding{55} & 362 & 167.0h & 53 & RGB, 3D Hand Pose \\
\hline
BRIO-TA \cite{BRIOTA} & 2022 & Assembly & Exo & 1 & 23 & MR, SR & PEs & 10 & \ding{55} & 75 & 2.9h & 15 & RGB \\
\hline
ATA~\cite{ATA} & 2023 & Toy Assembly/Lab & Exo & 3 & 15 & MR, SR & PEs & 11 & \ding{55} & 141 & 24.8h & 32 & RGB, Audio, Text, Pose, Obj. BB \\
\hline
HoloAssist \cite{HoloAssist} & 2023 & Assembly/Lab & Ego & 20 & 414 & MR, SR, ASD, SA, EMR & PEs, EEs & 16 & \ding{55} & 350 & 166h & 222 & RGB, Depth, Pose (Hand \& Head), Gaze, Text, IMU \\
\hline

CaptainCook4D \cite{CaptainCook4D} & 2023 & Cooking/Real & Ego & 24 & 352 & MR, SR, EMR, TAML & PEs, EEs & NA & \ding{55} & 384 & 94.5h & 8 & RGB, Depth, Text, Task Graphs \\
\hline
Ego-Exo4D\cite{grauman2024ego} & 2024 & Physical Exercise, Assembly/Real& Exo, Ego & 6 & 186 & MR, SR, SA & PEs & 5.6K$^{***}$ & \checkmark & 628 & 30h & 740 & RGB, Audio, Gaze, Depth, IMU, Text, Task Graphs, Bar, Mag \\
\hline
IndustReal\cite{IndustReal} & 2024 & Toy Assembly/Lab & Ego & 2  & 75 & MR, ASD, SR, TAML & PEs, EEs & 36 & \ding{55} & 84 & 5.8h & 27 & RGB, Depth, Gaze, Hand, Pose \\
\hline
EgoPER\cite{EgoPED} & 2024 & Cooking/Real & Ego & 5 & 70 & MR, TAML & PEs, EEs & 35 & \ding{55} & 386 & 28h & 11 & RGB, Depth, Audio, Gaze, Hand BB, Obj BB, Task Graphs\\
\hline
EgoOops\cite{EgoOops} & 2024 & Diverse/Real & Ego & 5 & 46 & MR, TAML & PEs, EEs & 58 & \ding{55} & 40 & 6.8h & 4 & RGB, Text, Object (label-only) \\
\hline
EK-M\cite{li2025mistake} & 2025 & Cooking/Real(synth.) & Ego & Inherited & 12.3K & MR, TAML & PEs, EEs & Inherited & \ding{55} & $>$221K & NATW & 37 & RGB, Text \\
\hline
Ego4D-M\cite{li2025mistake} & 2025 & Diverse/Real(synth.) & Ego &Inherited & 16K & MR, STAML & PEs, EEs & Inherited & \ding{55} & $>$257K & NATW & 248 & RGB, Text, PNR, BB \\
\hline
\end{tabular}
} 
\vspace*{0.10cm}
\caption{\textbf{Procedural understanding datasets targeting mistake detection.} \underline{Supported tasks:} MR: mistake recognition, ASD: activity state detection, SR: step (action) recognition, TAML: temporal action and mistake localization, STAML: spatiotemporal action and mistake localization, SA: skill assessment, EMR: early mistake recognition. Partps: Participants. Text refers to transcript availability. \underline{Note that}: TAML and STAML correspond to the general task of Mistake Detection (MD). \underline{Additional abbreviations}: PEs: procedural errors, EEs: execution errors, 3DM: 3D models publicly available, OF: Optical Flow, BB: bounding boxes, Bar: Barometer, Mag: Magnetometer. $^*$sparse annotations, $^{**}$ refined in subsequent works, $^{***}$ refers to total objects annotated across all dataset-supported tasks, only a subset used for mistake analysis (exact number not provided), NA: not annotated, NATW: non-available at the time of writing. \underline{Inherited}: the synthetic dataset inherits the specific attribute from the original.}
\label{tab:Datasets}
\end{table*}

\vspace{0.18cm}
\noindent \textbf{Size, modality diversity and domain generality:} Datasets like Assembly101 and EpicTent are, to this date, the largest in size, providing thousands of annotated sequences for tasks such as action recognition, anticipation and mistake detection. However, they tend to focus on a narrower set of tasks or domains (e.g., assembly or camping), which limits their domain generality. Ego-Exo4D, while smaller in comparison, offers valuable insights from both first-person and third-person perspectives, being the first dataset to incorporate synchronized egocentric and exocentric perspectives, enabling new opportunities for cross-perspective activity analysis, but lacks explicit mistake annotation. In contrast, datasets like CaptainCook4D and EgoOops are relatively smaller, but they offer a richer multi-modal information (e.g., depth, RGB, audio, gaze tracking), which makes them valuable for research in real-time and multi-modal systems. Nonetheless, large-scale datasets such as Assembly101 and ATA address industrial use cases and are focused on specific tasks, which may limit their generalization across other procedural activities. 
\vspace*{0.2cm}

\noindent \textbf{Hierarchical task structuring \& mistake taxonomies:} While several datasets, including EpicTent and CaptainCook4D, provide detailed hierarchical task and action step annotations, alignment between this task structuring and mistake annotations is often lacking. Several datasets offer valuable hierarchical annotations, providing insights into the decomposition of activities into sub-activities and action steps. EpicTent, CaptainCook4D, and IndustReal stand out, offering well-structured task steps that facilitate detailed action analysis within a broader task context. Assembly101, focused on assembly procedures, provides less granular annotations, primarily marking transitions between broader actions rather than detailed steps. Ego-Exo4D provides synchronized first- and third-person perspectives but lacks explicit hierarchical annotations, limiting support for fine-grained task structuring. Datasets like BRIO-TA, ATA, and CSV, while valuable in industrial settings, generally lack detailed hierarchical annotations, making them less suitable for in-depth hierarchical analysis. Holo-Assist offers some task segmentation but lacks the hierarchical breakdown found in EpicTent and CaptainCook4D. Finally, EgoOops, though more focused on mistake detection than datasets such as Assembly101, does not emphasize task structuring, hindering efforts to understand the linkage between mistake occurrence and task decomposition.

Regarding the mistake annotation specificity and the presence of clear taxonomies, these aspects remain highly inconsistent across datasets. Existing datasets vary in the tasks they cover and the error types they capture. IndustReal, CaptainCook4D, and EgoOops provide rich, context-sensitive annotations of both execution and procedural errors, such as incorrect tool use, missing steps, or mis-sequenced actions, particularly in egocentric settings. Epic-Tent, BRIO-TA, and ATA focus more on task steps, action sequences, and task verification, with annotations centered on procedural errors like skipped or misordered steps, thus remaining task-specific rather than context-sensitive. Conversely, Assembly101, HoloAssist, and Ego-Exo4D were designed for procedural activity understanding, with mistake recognition only a downstream task. Consequently, no well-defined error taxonomy exists, and error annotations remain sparse or implicit.
\vspace*{0.05cm}

\noindent \textbf{Mistake/error accumulation:}  Most of the reviewed datasets, including Assembly101 (original), EpicTent, CaptainCook4D, BRIO-TA, ATA, CSV, EgoOops and Ego-Exo4D, do not explicitly model or annotate error accumulation. For instance, in furniture manufacturing, a missing screw may lead to the incorrect placement of another component, which in turn could compromise the structural integrity of the final assembly, potentially resulting in collapse. In such cases, an erroneous prior state can propagate additional errors, exacerbating the overall failure. In many of these datasets, errors are either isolated events or are followed by corrections that resolve the problem within the same sequence segment. 
For example, Assembly101 (original) and EpicTent include procedural errors but lack temporal linking that would allow tracking their downstream impact. Similarly, EgoOops emphasizes error detection but not how one error might lead to others. CaptainCook4D offers multimodal recordings with rich supervision but treats errors at a segment level without modeling cumulative consequences.

Exceptions are the variant of Assembly101 proposed by Ding et al.~\cite{ding2023mistakecountsassembly} and, more recently, the IndustReal dataset. The first incorporates the notion of error accumulation in the mistake detection task using a rule-based action transition analysis scheme, where an action transition rule is classified as transitive or intransitive, depending on how mistakes propagate through the action sequence. Transitive rules imply that any subsequent dependent action after an incorrect anchor action is also a mistake, while intransitive rules consider only specific dependent actions as mistakes. Although very generalizable, this approach is accompanied by physical annotations of specific cases in the dataset. Contrary, IndustReal emphasizes long-horizon task monitoring in real industrial environments, where an initial error can propagate through subsequent actions. Its design supports tracing such error chains, offering a realistic depiction of procedural degradation over time, crucial for systems aiming to assist real-world task execution.
\section{Evaluation Metrics in Mistake Analysis}

The evaluation of models for mistake analysis in procedural activities is tightly coupled with the task formulation, which includes identifying a mistake (mistake recognition), anticipating it before it fully unfolds (early mistake recognition), or temporally localizing it within a video sequence (mistake detection). In this section, we categorize evaluation protocols in existing works by target task and further divide them into two groups: (a) \textit{generalized metrics}, assessing overall system performance using standard classification or detection scores, and (b) \textit{error type-specific metrics}, evaluating performance over specific mistake categories or structural aspects, such as sequencing violations or execution errors.

\subsection{Generalized Metrics}
\subsubsection{Mistake Recognition}
The predominant evaluation strategy adopted in prior work~\cite{EgoPED} involves formulating mistake recognition either as a \textit{binary classification task}, wherein each action is labeled as correct or mistaken, or as a \textit{multi-class classification task} that further distinguishes between specific error types. In these settings, standard classification metrics such as precision, recall, F1 score and the area under the receiver operating characteristic curve (AUC) are commonly used to quantify model performance.

While these general-purpose metrics provide a foundational assessment, several works have proposed adaptations that better reflect the temporal and procedural nature of the mistake detection task. One such task-specific metric is the \textit{Error Detection Accuracy (EDA)}~\cite{EgoPED,kung2025changed}, originally introduced in~\cite{EgoPED}. EDA is defined as the ratio of correctly detected erroneous segments over the total number of ground-truth erroneous segments across all test videos. segments. Specifically, given the test set, $V_t$, consisting of $K$ videos, $V_t = \{ V_1, \dots, V_K \}$, where each video $V_k$ is decomposed into a set of segments $V_k = \{ v_i^{(k)} \}_{i=1}^{N_k}$, with corresponding ground-truth mistake labels $m_i^{(k)} \in \{0,1\}$ and predicted mistake labels $\hat{m}_i^{(k)} \in \{0,1\}$, the EDA is defined as:
\begin{equation}
    \mathrm{EDA} = 
    \frac{
        \sum_{k=1}^{K} \sum_{i=1}^{N_k} \mathbf{1}\{ m_i^{(k)} = 1 \wedge \hat{m}_i^{(k)} = 1 \}
    }{
        \sum_{k=1}^{K} \sum_{i=1}^{N_k} \mathbf{1}\{ m_i^{(k)} = 1 \}
    },
\end{equation}
\noindent
where $\mathbf{1}\{\cdot\}$ denotes the indicator function, returning $1$ if its argument is true and $0$ otherwise. A segment $v_i^{(k)}$ is considered erroneous ($m_i^{(k)} = 1$) if at least a subset of its frames is labeled as a mistake, following a frame-centered evaluation scheme. This formulation aggregates detections across all test videos and captures the model’s ability to correctly localize error-prone intervals while tolerating limited temporal imprecision at segment boundaries. Essentially, EDA is a \textit{recall-type} metric that quantifies the proportion of ground-truth erroneous segments successfully recognized across the entire test set.

\vspace*{0.10cm}

\noindent A related metric, the \textit{Anomaly Segment Accuracy (ASA)}, was introduced in the BRIO-TA dataset~\cite{BRIOTA} to quantify segment-level performance in mistake recognition and detection tasks. ASA is a segment-level analogue of the standard accuracy metric, adapted to mistake analysis. The ASA metric measures the proportion of action segments that are correctly classified as either normal or erroneous, and is defined as
\begin{equation}
    \mathrm{ASA} = 
    \frac{
        \sum_{k=1}^{K} \sum_{i=1}^{N_k} \mathbf{1}\{\hat{m}_i^{(k)} = m_i^{(k)}\}
    }{
        \sum_{k=1}^{K} N_k
    },
\end{equation}
\noindent
where $\mathbf{1}\{\cdot\}$ denotes the indicator function. 

By construction, ASA treats both correct and erroneous segments symmetrically, in contrast to EDA which focuses exclusively on erroneous segments. Consequently, ASA provides a balanced segment-level measure of classification performance, assessing correctness across the entire procedural execution rather than solely focusing on erroneous segment localization.

\subsubsection{Early Mistake Recognition}
This task resembles early action recognition, where the goal is to anticipate an event before it fully unfolds. Therefore, similar protocols and evaluation metrics have been partially adopted. In early action recognition~\cite{zhao2021review}, common strategies include measuring classification accuracy at fixed observation percentages (e.g., $10\%$, $20\%$, …, $100\%$ of the action duration, termed \textit{anticipation windows}) or computing metrics like AUC to capture performance over time. Similarly, recent works in early mistake recognition~\cite{CaptainCook4D}, \cite{CSV}, \cite{HoloAssist} evaluated model performance at different anticipation windows, gauging predictive accuracy before the full mistake occurred. As an example, CaptainCook4D\cite{CaptainCook4D} introduces an early error detection task, where models are evaluated at fixed intervals prior to the annotated mistake point (e.g., 2s before). Performance is measured using Top-1 accuracy and F1-score within these anticipation windows. This allows for assessing how well the model can detect an impending mistake with limited temporal context. In a related but slightly different fashion, SVIP~\cite{SVIP} evaluates sequence verification models on partially observed video snippets to assess the ability to recognize incomplete or incorrect procedures early on. While SVIP focuses on binary classification accuracy of a completed vs. incomplete (or incorrect) procedure, its use of early segment evaluation reflects similar motivations to those found in early action recognition literature.

Although these works do not always explicitly refer the term anticipation window, they embody the same principle: evaluate how reliably a model flags mistakes before they fully manifest.

\subsubsection{Mistake Detection}
As outlined in Section~\ref{MistakeTasks}, mistake detection can be formulated as a \textit{temporal action localization} (TAL) problem, where the objective is to identify the type of mistake and its temporal boundaries within the video. Under this format, several recent works~\cite{EgoOops}, \cite{CaptainCook4D, BRIOTA} adopt evaluation protocols derived from established TAL benchmarks~\cite{lai2024human, zhong2023survey}, employing metrics such as mean Average Precision (mAP) and Recall at X (R@X). These metrics are computed at varying temporal Intersection-over-Union (tIoU) thresholds, typically ranging from 0.1 to 0.5 in increments of 0.1 or 0.2.

Under this scheme, mAP is calculated across all mistake action classes, explicitly excluding the ``correct'' class to isolate error detection performance. Furthermore, a predicted mistake label is valid only if it corresponds to the same procedural step as the ground-truth segment, thereby enforcing alignment in both temporal and semantic dimensions. This evaluation strategy ensures that models are assessed not only for their ability to detect mistakes but also for their step-level precision.

Beyond conventional TAL metrics, some recent works have introduced task-specific measures that capture higher-order aspects of procedural consistency. Notably, Schoonbeek et al.~\cite{IndustReal} propose the \textit{Procedure Order Similarity (POS)} metric to assess how well the predicted sequence of completed steps adheres to the canonical step order. POS is defined as:

\begin{equation}
    \text{POS} = 1 - \min\left( \frac{\text{DamLev}(\mathcal{P}, \hat{\mathcal{P}})}{|\mathcal{P}|},\ 1 \right),
\end{equation}

\noindent
where $ \mathcal{P} = \{ a_i\}_{i=1}^N$ is the ordered ground-truth step sequence, $\hat{\mathcal{P}} = \{ \hat{a}_j\}_{j=1}^M$  is the predicted sequence,  and \( \text{DamLev} \) denotes the Damerau–Levenshtein distance~\cite{damerau1964technique} without substitutions. This metric accounts for missing, repeated, or wrongly ordered steps providing a holistic measure of procedural correctness.

\subsection{Mistake/Error Type-specific Metrics}
\noindent \textbf{Omission errors using action frequencies:} The work of Ghoddoosian et al.~\cite{ATA} proposed a metric to evaluate the performance of mistake detection methods by expressing the discrepancies between expected and predicted action frequencies for specific action pairs. The metric relies on defining error-specific functions $\{F_e\}$ that operate over these frequencies. Each function maps the observed frequency $f_a$ of actions in the test video to the number of instances $n_e$ in which an error $e$ occurs. For example, the error\textit{ Loose Assembly}, arising when a component is positioned but not secured, is expressed as:
\begin{equation}
F_{\text{Loose Assembly}} = \max(f_{\text{place component}} - f_{\text{tighten bolt}}, 0),
\end{equation}
where $f_a$ denotes the frequency of action $a \in \{\text{place component}, \text{tighten bolt}\}$ within the video sequence.

To evaluate mistake detection, a baseline method is typically employed, which involves generating two distinct segmentation results: $S_0$ (a constrained offline step segmentation using a reference transcript from the training set) and $S_{\tau}$ (a predicted step segmentation with $\tau > 0$). The corresponding action frequency distributions, $f_0$ and $f_{\tau}$, are computed for these segmentations. Using these distributions, the occurrence of each error $e$ is determined as:
\begin{equation}
n_e = F_e(f_{\tau}) \times (1 - \min(F_e(f_0), 1)).
\end{equation}

This formulation ensures that errors are detected based on deviations from the reference segmentation while mitigating false positives. Specifically, the term $(1 - \min(F_e(f_0), 1))$ conditions error detection on discrepancies between the predicted and reference transcripts. The computed error occurrences $n_e$ are subsequently used to quantify the overall error detection performance by computing the F1-score. \vspace{0.1cm}

\noindent \textbf{Omission errors using edit distance:} Lee et al.~\cite{EgoPED} introduced the Omission Intersection over Union (O-IoU) metric to evaluate the performance of mistake detection methods, particularly in scenarios involving omissions, defined as:

\begin{equation}
\text{O-IoU} = \frac{|GT_o \cap D_o|}{|GT_o \cup D_o|},
\end{equation}

\noindent where \(\{ GT_o, D_o\} \) refer to the set of ground-truth, and detected omission errors respectively. The set \( D_o \) is determined by identifying the closest step sequence from the training videos to the predicted steps in the test video, based on the Edit distance. Specifically, given $\hat{\mathcal{P}}$, the set of predicted steps (actions) derived from the an action segmentation method, and $\mathcal{P}$ the set of steps corresponding to the best-matched training sequence, then, the set of omitted steps, \( D_o \), is estimated as the ratio \( D_o = { \mathcal{P}}/{ \hat{\mathcal{P}}} \), i.e. all steps in $\mathcal{P}$ missing from $\hat{\mathcal{P}}$. \vspace{0.1cm}

\noindent\textbf{Procedural error assessment using Weighted Distance Ratio (WDR):} Qian et al.~\cite{CSV} introduced WDR as an evaluation metric for procedural mistake detection, which quantifies the performance of a method by assessing the similarity between action sequences from correct and incorrect procedural executions. WDR is defined as the ratio of the average distance between negative video pairs to the average distance between positive video pairs. A \textit{positive pair} consists of two videos in which the procedural steps are executed correctly and in the correct order, whereas a \textit{negative pair} contains at least one video with a procedural error. The metric is defined as:
\begin{equation} 
WDR = \frac{\frac{1}{N} \sum_{i=1}^{N} wd_i}{\frac{1}{P} \sum_{j=1}^{P} d_j},
\end{equation}
where \(P\) and \(N\) denote the number of positive and negative pairs, respectively. Here, \(d_j\) represents the Euclidean distance between the learned video embeddings in a negative video pair \((V_{j,1}, V_{j,2})\). In contrast, \(wd_i\) is the \textit{normalized distance} for a positive video pair \((V_{i,1}, V_{i,2})\), defined as
\(
wd_i = d_i/{ed_i},
\)
where \(d_i\) is the Euclidean distance between the learned embeddings of the videos, and \(ed_i\) is the Levenshtein distance~ between the corresponding text sequences of pair \(i\). The text sequences symbolically represent the procedural steps performed in the video (e.g., ``pick part → attach part → tighten screw''), enabling the metric to account for semantic similarity at the step level in addition to visual similarity.

A higher $WDR$ indicates better performance, as it reflects a model’s ability to clearly separate incorrect sequences (larger distances) from correct ones (smaller distances). High $WDR$ suggests that a model effectively distinguishes between valid and invalid sequences. Conversely, a low WDR indicates poor differentiation between correct and incorrect sequences.

\subsection{Mistake Explanation Metrics}
With the emergence of methods that provide natural language justifications for detected errors~\cite{patsch2025mistsense}, evaluating the semantic quality and diagnostic accuracy of these explanations has become a critical sub-task. Current approaches primarily repurpose standard image and video captioning metrics to assess the similarity between the generated explanation and a ground-truth reference.

The most common evaluation protocols utilize N-gram overlap metrics, including BLEU\cite{papineni2002bleu}, {ROUGE}\cite{lin-2004-rouge}, and {CIDEr}\cite{vedantam2015cider}. These metrics quantify performance by calculating the precision and recall of word sequences between the predicted and reference explanations. For instance, Patsch et al.\cite{patsch2025mistsense} employ CIDEr as a primary metric to capture the consensus of generated explanations with human annotations. While these metrics effectively measure linguistic fluency and textual similarity, they can fail to capture \textit{diagnostic correctness} in the context of mistake analysis. A generated explanation could achieve a high BLEU score by matching the majority of a sentence (e.g., ``\textit{The user failed to pick up the object}'') while missing or hallucinating the critical causal detail (e.g., specifying the wrong object).

\section{Related Work on Mistake Analysis Tasks}
\label{MistakeTasks}

In this section, we provide a comprehensive overview of mistake analysis tasks and review the existing literature. A broader categorization and presentation of methods, organized according to their methodological characteristics rather than the specific tasks they target, is presented in Section~\ref{sec: method_grouping}. It is important to note that, for mistake recognition, the majority of existing studies treat it as part of the broader mistake detection problem. Accordingly, our analysis focuses on recognition within the context of detection frameworks. Additionally, in these works the classification of error types is typically formulated as a downstream classification task.


\subsection{Mistake Detection and Recognition}
As stated in Section~\ref{DefMis}, mistake detection operates directly on raw video, requiring both error localization and classification, though in existing approaches this localization is inherently constrained to the action level, aligning with predefined boundaries. Consequently, methods can only determine whether an entire action instance contains an error, overlapping with mistake recognition, which assumes access to semantically labeled clips. This limitation stems from the lack of fine-grained temporal localization, pinpointing the precise moment of error occurrence within action boundaries,in most datasets. We discuss these challenges further in Section~\ref{chal_probs}.

Several works~\cite{CaptainCook4D,EgoOops,ATA} tackled mistake detection as a temporal action localization problem, assuming that deviations from expected procedural flow, such as missing steps, incorrect orderings, or executional anomalies, manifest as misaligned or inconsistent temporal patterns. Most approaches build upon self- or fully supervised action segmentation models fine-tuned on task-specific action pools to segment untrimmed videos into procedural steps. Mistake presence is then inferred through misalignment with expected sequences, unexpected transitions, or out-of-distribution behavior. Methods such as~\cite{CaptainCook4D,EgoPED} employ spatiotemporal backbones (I3D~\cite{carreira2017quo}, SlowFast~\cite{feichtenhofer2019slowfast}) with segmentation or detection modules (e.g., ActionFormer~\cite{zhang2022actionformer}, MS-TCN++~\cite{farha2019ms}, MiniRoad~\cite{an2023miniroad}) followed by mistake recognition processes like segment-level classification~\cite{CaptainCook4D}, misalignment scoring~\cite{EgoPED}, or temporal inconsistency detection~\cite{EgoOops}. Some also exploit text-based procedural priors alongside RGB to identify temporal inconsistencies, for example, EgoOops~\cite{EgoOops} aligns predicted actions with task descriptions to highlight erroneous segments. In weakly supervised settings~\cite{ATA}, models rely solely on ordered transcripts and constrained alignment objectives to localize unseen mistakes without direct annotations. Despite architectural differences, these methods share a common goal: to detect \textit{when} mistakes occur in untrimmed video and to \textit{segment} the corresponding intervals.

A recent work by Li et al.~\cite{li2025mistake} pushes the boundaries of offline mistake detection beyond simple temporal intervals by introducing \textit{Mistake Attribution (MATT)}. Unlike standard detection schemes followed by existing methods, which flag the entire duration of a deviation (entire action segment (procedural error) or a portion of the action segment (executional error)), MATT aims to spatiotemporally localize the mistake occurrence. To achieve this, they introduce a mechanism to detect the Point-of-No-Return (PNR) frame, which corresponds to the exact instant the mistake becomes irreversible. The PNR serves as a temporal anchor for spatial reasoning, enabling models to localize the mistake's visual manifestation (e.g., via bounding boxes) on the exact frame where the error consolidates.

Finally, recent works attempt to reformulate this offline setting into an online pipeline, making predictions incrementally as new frames arrive, using only past context. Flaborea et al.~\cite{flaborea2024prego} introduced PREGO, combining an \textit{online action recognition} module with a \textit{symbolic reasoning} branch driven by a pre-trained LLM that predicts the next expected action and flags discrepancies as mistakes. Plini et al.~\cite{plini2024ti} extend this by adding chain-of-thought reasoning and few-shot in-context examples, improving adaptability and temporal precision.

\subsection{Early Mistake Recognition}

\textcolor{black}{Early mistake recognition methods aim to identify mistakes in video segments when only an initial portion of a procedural step is observed. Unlike early action recognition~\cite{kong2022human}, which seeks to infer the action being performed from a predefined action set, early mistake recognition focuses on detecting deviations from the intended action based on partial observations. This is particularly challenging as it requires attending subtle cues in the evolving action dynamics, while mistakes may manifest in diverse and often previously unseen forms.}

Despite this strong relationship between the two tasks and the extensive research dedicated to the former, the latter has received comparatively less attention, with only a few works addressing it~\cite{CaptainCook4D}, \cite{CSV}, \cite{HoloAssist}. Qian et al.~\cite{CSV} target mistake detection and early mistake recognition as a video alignment problem, leveraging a Transformer-based architecture to align pairs of video sequences while jointly optimizing a video-level activity recognition loss. To facilitate early mistake recognition, they propose a baseline method based on the assumption that a reference, correct execution of an activity, \( P_{\text{ref}} \), and a candidate execution, \( P_{\text{cand}} \), which may contain mistakes, generally span similar time intervals. Under this assumption, the time span of both executions is partitioned into \( k \) intervals and the \( \ell_2 \) distance between \( P_{\text{ref}} \) and \( P_{\text{cand}} \) is computed in the feature space \( f \). A sudden increase in the similarity distance acts as an indicator of an unexpected event, which may correspond to a mistake in the execution process.

Wang et al., in HoloAssist~\cite{HoloAssist}, introduce a formulation that indirectly supports early mistake recognition via an \textit{intervention forecasting task}. Framed as a temporal prediction problem, the model anticipates the need and type of intervention based on observed task progression. Interventions were delivered verbally by a domain expert monitoring the actor’s performance in real time. These serve as supervisory signals, marking critical moments where deviations from expected task execution warrant external assistance, enabling models to learn patterns indicative of upcoming failures. Finally, Peddi et al.~\cite{CaptainCook4D}, in \textit{CaptainCook4D}, formulate early mistake detection in alignment with classical action anticipation paradigms~\cite{lai2024human}, where a model receives a partial observation of an action segment and predicts whether it will result in a wrong or correct execution. They benchmarked this task with top-performing convolutional (SlowFast~\cite{feichtenhofer2019slowfast}, X3D~\cite{feichtenhofer2020x3d}) and transformer-based models (Omnivore~\cite{girdhar2022omnivore}, VideoMAE~\cite{tong2022videomae}) by giving them access to only the first half of an action segment. Experiments showed a significant performance drop over full-segment mistake recognition, highlighting the increased difficulty of forecasting errors under limited temporal context.

\section{Methodological and Supervision-Based Grouping of Mistake Analysis Approaches}
\label{sec: method_grouping}

Procedural activities are inherently structured, with each action contributing to a defined sequence toward a specific goal. Beyond this sequential organization, the internal characteristics of each action instance, i.e. its constituent steps, object manipulations, and evolving scene states, also determine the activity execution correctness. These intra-action dynamics often reveal subtle deviations that may not disrupt the global sequence but still signify incorrect or suboptimal execution, particularly in executional error analysis. Comprehensive mistake detection thus requires reasoning about correctness both at the macro sequence level and the micro patterns within actions. This process typically follows a multi-stage pipeline, where low-level visual perception modules (e.g., object detection~\cite{ObjSurvey}, pose~\cite{PoseSurvey} and gaze estimation~\cite{cazzato2020look}) extract cues from video input. Mid-level modules then track interactions, model temporal dynamics, or recognize ongoing actions. Finally, high-level reasoning components compare extracted information against predefined task models or learned activity graphs to assess alignment with expected execution patterns. Deviations at the action or sequence level are flagged as potential errors.

Existing methods for mistake analysis often share core design principles, i.e. video inputs, temporal modeling, and action understanding, but differ in problem formulation, supervision use, and procedural knowledge integration. The broadest categorization follows the overarching design pipeline, shaped by the targeted error class: procedural, executional, or both. Procedural-focused methods emphasize step dependencies, action ordering~\cite{seminara2024differentiable},\cite{plini2024ti}, or task progression models~\cite{CSV}, whereas executional ones~\cite{mazzamuto2025gazing} focus on fine-grained motion, posture, or object manipulations, often ignoring broader context. Hybrid approaches~\cite{EgoPED} attempt to unify both perspectives, typically requiring more complex architectures capable of jointly modeling high-level task structure and low-level signals. The mistake family a method targets influences its learning objective, supervision level, and integrated priors. We group existing methods along four key pillars: {(a)} the specific mistake task; {(b)} supervision regime; {(c)} learning paradigm; and {(d)} degree of procedural structure integration, from structure-agnostic to hierarchy-aware models. Table~\ref{tab:method-taxonomy} summarizes representative methods along these dimensions.
\subsection{Procedural Structure Utilization}

\begin{figure*}[t]
    \centering
    \includegraphics[width=0.95\linewidth]{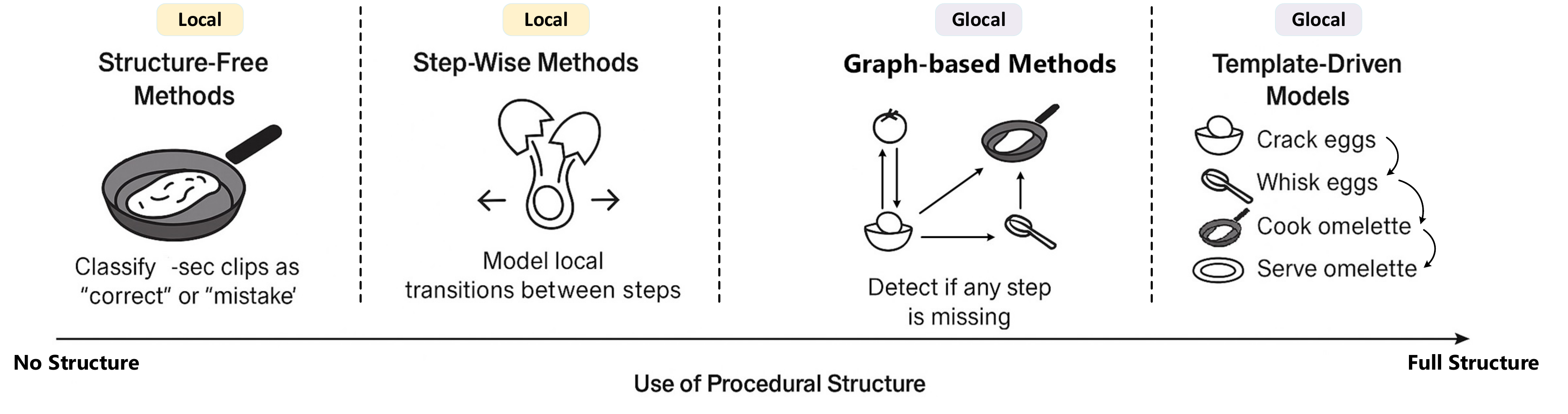}
        \caption{Method taxonomy based on procedural structure usage. Horizontal axis reflects the \textit{degree of structure imposed}, from flat models that predict without procedural priors to structured ones that explicitly encode or reference task structure. Vertical axis captures the \textit{scope of structure}, with local scope models learning short-range temporal dependencies and global scope models being capable to reason over the full procedural execution. Sketches illustrated with \cite{openai_chatgpt}.
}

    \label{fig:proc_structure}
\end{figure*}

Understanding and analyzing multi-action activities from video data requires models to capture temporal structure, action-wise dependencies and deviations from ideal procedures. As showcased in the previous section, recent research in procedural video understanding has introduced a variety of modeling paradigms that differ in how they represent and utilize procedural knowledge. Under the premise of how (and whether) a mistake analysis method incorporates knowledge of the procedure, we can categorize existing methods into four broad classes based on the degree of structure imposed and the source of procedural guidance: \textit{structure-free}, \textit{step-wise}, \textit{graph-based} and \textit{template-driven} models. A visual overview of the categorization scheme based on the scope and utilization degree of the procedural knowledge is shown in Figure~\ref{fig:proc_structure}.

\subsubsection{Structure-Free Methods} 

Structure-free methods (SFM) approach procedural video understanding without imposing any prior assumptions about the underlying task structure. These models treat the input video as a continuous or disjoint sequence of frames or clips, learning to predict action labels directly. While simple and often scalable, they lack an explicit notion of step boundaries or temporal dependencies, which can limit their ability to capture procedural coherence or detect structural anomalies such as missing or disordered steps. Peddi et al.~\cite{CaptainCook4D} evaluated several baseline methods in  CaptainCook4D, focusing on structure-free approaches for supervised mistake recognition. Specifically, they examined the performance of VideoMAE~\cite{tong2022videomae} with a linear classifier, X3D~\cite{feichtenhofer2020x3d}, Omnivore~\cite{girdhar2022omnivore} and SlowFast~\cite{feichtenhofer2019slowfast} models. These frameworks classify short video clips as correct/mistake using only visual features from fixed-length segments, without leveraging procedural context, sequential dependencies, graph-based reasoning, or alignment with predefined task templates.

Under a different setting, Lee et al.~\cite{EgoPED} utilize a two-stage, structure-free framework for procedural error detection. A temporal convolutional network first segments the input into per-frame step assignments, augmented with relational hand/object graphs to generate rich feature representations. For each step, a set of \emph{prototypes} of correct executions is learned via contrastive loss, pulling frame features toward nearest prototypes and pushing them from others. At test time, each frame is scored by its cosine similarity to the closest prototype of the predicted step; low similarity indicates a procedural error (omission, addition, modification, slip, or correction).

A distinct line of research within the structure-free paradigm is proposed by Mazzamuto et al.~\cite{mazzamuto2025gazing}, who approach mistake detection through unsupervised gaze modeling. Rather than relying on annotated action steps, temporal boundaries, or an explicit procedural structure, their method treats gaze behavior as an implicit signal of task regularity. They introduce a gaze completion task, training a model to predict future gaze distributions from past egocentric video and eye movements during correct task executions. Mistakes are detected by comparing predicted and observed gaze, with significant deviations indicating potential attentional inconsistencies in execution.

\subsubsection{Step-wise Models} 
Step-wise models (STM) explicitly incorporate procedural structure by segmenting a task into discrete steps or actions, without the use of an explicit task graph (no nodes‐and‐edges encoding of step dependencies) and no template alignment against a master protocol. 

The foundational methods in this category typically follow a supervised two-stage pipeline: first segmenting the action, then classifying it as correct or incorrect. Wang et al.~\cite{HoloAssist} utilize a pre-trained ViT~\cite{dosovitskiy2020image} backbone to encode video clips corresponding to annotated action steps and feed the resulting per-step embeddings into a TimeSformer~\cite{bertasius2021space} coupled with a classification head that jointly predicts the step label and a binary “correct vs. mistake” flag. Similarly, the baseline provided in the BRIO-TA dataset~\cite{BRIOTA} utilizes spatiotemporal backbones like I3D~\cite{carreira2017quo} or X3D~\cite{feichtenhofer2020x3d} followed by temporal segmentation models (e.g., MS-TCN~\cite{farha2019ms}) to output frame-wise mistake predictions.
Advancing this paradigm, Patsch et al.~\cite{patsch2025mistsense} introduce MistSense, which fuses explicit hand pose features with RGB data, each modelled via a dedicated temporal encoder and aligned via Q-Formers, to target fine-grained \textit{executional errors}.

While these approaches treat the step as a monolithic unit for binary classification, recent work by Li et al.~\cite{li2025mistake}  advances the STM paradigm via \textit{Mistake Attribution}. Instead of a simple binary flag, their method conditions the step-wise analysis on Semantic Role Labeling~\cite{gardner2018allennlp}, decomposing the textual description of an action into specific components (Predicate, Object). By cross-attending these role tokens with the video step features, the model can explicitly attribute the error to a specific semantic violation (e.g., recognizing that the \textit{object} was manipulated incorrectly despite the correct \textit{action} being performed). This moves modeling from coarse detection to fine-grained semantic and spatiotemporal localization.

A notable departure from these supervised paradigms is the approach  in~\cite{EgoPED}, which despite following a similar action localization pipeline as previous methods, it introduces a Contrastive Step Prototype Learning (CSPL) mechanism enabling step-wise mistake detection in an unsupervised manner. Instead of learning directly from both correct and erroneous executions, CSPL exploits only correct sequences. The model learns discriminative step-specific prototypes in a contrastive embedding space (via InfoNCE loss~\cite{oord2018representation}), encouraging the network to produce similar embeddings for instances of the same step while pushing away different ones. At inference, each step in a test sequence is projected into this space and classified based on proximity to its corresponding prototype. Deviations from the expected prototype are flagged as errors. 

Finally, a recent step-wise approach is the PECC framework~\cite{ICME}, which extends the two-stage paradigm with a probabilistic treatment of step embeddings. As in prior STM models, PECC first applies a TAS backbone to obtain per-frame step labels, enforcing an explicit action segmentation. A Causal Dilated Convolution (CDC) module refines TAS features to preserve temporal dependencies and ensure causal consistency across boundaries. For each segmented step, PECC fits a Gaussian Mixture Model (GMM) over normal-execution features, yielding a probabilistic representation of step-specific appearance and motion. During inference, frames are scored by per-step log-likelihood under the corresponding GMM and flagged as errors when their likelihood deviates from the learned distribution. While it reliably identifies execution errors, it captures procedural ones only indirectly, typically when misordering or omissions cause inconsistent TAS predictions. Like all models in this category, its performance remains tied to the quality of the underlying segmentation.
\subsubsection{Graph-Based Models} 

Graph-based approaches represent procedural knowledge as structured graphs, where nodes correspond to actions and edges encode transitions or dependencies. These models leverage the graph structure to reason over the procedural space, enabling more robust inference of task progression and detection of deviations. This method family is the most prominent in mistake analysis, with existing works being grouped into three distinct clusters based on the type of knowledge embedded in the graph and the way structure is enforced or utilized: \textit{(i)~procedural task graphs}, \textit{(ii) semantic knowledge graphs} and \textit{(iii) rule-based graphs}.
\vspace*{0.1cm}

\noindent\textbf{Procedural Task Graph methods (PTG):} These methods utilize acyclic graph structures to explicitly encode the temporal and logical structure of a procedure, where nodes typically represent discrete action steps or intermediate scene states, and, edges define valid transitions or causal dependencies. Such graphs may be manually annotated or automatically induced from video data~\cite{seminara2024differentiable}. They are particularly effective in modeling normal task flow and detecting anomalies, as mistake events often correspond to deviations from expected graph transitions. 

Seminara et al.~\cite{seminara2024differentiable}, in a seminal work, investigated both approaches and their impact on the action understanding problem and the downstream task of mistake detection. For the more challenging variant of automatic graph construction, they proposed a differential learning scheme, where a neural encoder extracts frame-level features, while a probabilistic adjacency matrix, jointly optimized with step recognition, models step dependencies. To enable differentiable graph learning they proposed the Task Graph Maximum Likelihood (TGML) loss, that allows end-to-end optimization of the task graph's adjacency matrix. It comprises two core components: (a) \textit{a positive reinforcement term}, maximizing edge probabilities in the adjacency matrix for observed action transitions and (b) \textit{a contrastive penalization term}, penalizing edges corresponding to action transitions absent in training sequences.

Building on the use of explicit procedural structures, Lee and Elhamifar~\cite{lee2025error} propose \textit{GTG2Vid}, a method that formalizes the alignment of an observed video sequence to a predefined  task graph as a dynamic programming problem. The method defines novel frame and node skipping dropping policies via a cumulative cost function over frame-node assignments, combining: (i) \textit{matching costs} between each frame and candidate graph node (action), (ii) \textit{frame-drop costs} that penalize advancing in time without transitioning to a new graph node, allowing the model to ignore ambiguous or visually uninformative frames (background actions) and (iii) \textit{node-drop costs} that penalize skipping graph nodes, enabling the representation of omitted, reordered, or prematurely bypassed procedural steps.  On top of this alignment, the \textit{Error Recognition Module} evaluates the visual consistency of each frame with the expected embedding of its assigned graph node, identifying off-graph behavior as mistakes.  

While previous methods optimize for alignment to a single path, Huang et al.~\cite{huang2025modeling} leverage the task graph to explicitly handle execution variability.  In their AMNAR framework, the graph functions as a query engine that retrieves \textit{all} permissible transitions given the current state, rather than enforcing a unique trajectory. This allows the model to generate adaptive "normal" representations for multiple valid branches, effectively distinguishing between graph-compliant variations and true off-graph mistakes.
\vspace{0.06cm}

\noindent\textbf{Semantic Knowledge Graph methods  (SKG):} These methods utilize graphs to represent semantic relationships among objects, actions and their attributes within the task domain. These graphs are constructed from textual knowledge bases, external corpora, or structured ontologies. Rather than focusing on temporal ordering, semantic graphs provide contextual grounding for actions by representing affordances, object-action interactions and high-level concepts. When combined with vision models, they support reasoning about action plausibility given the current scene configuration, offering indirect support for error detection via semantic inconsistency analysis. Despite the potential of this strategy to capture richer affordance-level reasoning and to enforce semantic validity of action–object interactions, existing methods have not widely adopted it due to practical challenges: fine-grained objects in procedural tasks are difficult to detect and track reliably, generic detectors rarely cover task-specific components, and grounding semantic relations in visual representations requires substantial task-specific annotation and model fine-tuning. These limitations make robust semantic reasoning difficult to achieve in real-world procedural video settings.
\vspace*{-0.05cm}

\noindent \textbf{Rule-based Graph methods (RG):} These methods incorporate domain knowledge in the form of explicit procedural rules or logic, often defined by experts or extracted from formal manuals and usually are combined with the former two classes. The rules govern permissible action transitions or constraints on execution order, giving rise to graphs that represent normative task flows. Such representations are well-suited to high-precision domains such as industrial assembly or medical workflows, where deviations from the rule-defined paths are critical indicators of mistakes.  Although each cluster reflects a different design philosophy for incorporating prior knowledge and modeling task regularities, a number of existing methods can belong to more than one class.  An exemplar hybrid approach is the seminal work of Ding et al.~\cite{ding2023mistakecountsassembly}, build a knowledge base of spatial (object-to-object relations) and temporal (orderings of spatial relations) graph structures and defines a set of logical rules in the temporal structure to detect ordering mistakes in the toy assembly and disassembly procedures in the Assembly101~\cite{Assembly101} dataset.
\vspace{0.03cm}

\noindent \textbf{Template-Driven Model (TDM):} TDMs incorporate \textit{external procedural templates}, such as transcripts, instructional scripts, or curated step sequences, as priors to guide the understanding of complex multi-step activities. These templates define the \textit{canonical action ordering} and serve as reference structures against which observed video sequences are aligned. Essentially, given a procedural template represented as an ordered action sequence and the predicted sequence, TDMs seek the optimal alignment between them under temporal consistency or monotonicity constraints, interpreting deviations as procedural or executional errors. These models leverage strong task priors encoding \textit{what should be done and in what order}, enabling both action segmentation and mistake detection under weak supervision.  Ghoddoosian et al.~\cite{ATA} employ a method based on action transcripts, using the \textit{Constrained Discriminative Forward Loss (CDFL)}~\cite{li2019weakly} to align frame-level predictions with expected transcripts while maximizing the decision margin between valid-invalid labels. Mistake detection is set by measuring deviations from the canonical template without error examples during training. Narasimhan et al.~\cite{narasimhan2023learning} introduce \textit{VideoTaskformer}, which learns procedural structure by predicting randomly masked steps in instructional videos. Instead of aligning predictions to transcripts, it applies a masked modeling objective~\cite{devlin2019bert} over ordered step descriptions, using a transformer to reason over global context and identify missing or out-of-order steps disrupting procedural flow. Schoonbeek et al.~\cite{IndustReal} employ a two-stage template-driven pipeline: an Assembly State Detection (ASD) module based on YOLOv8-m~\cite{yolov8_ultralytics} recognizes current assembly states from video, followed by a Procedure Step Recognition (PSR) stage mapping detected states to step sequences via non-parametric heuristics. Deviations from canonical templates, defined by task instructions, indicate procedural error presence.

\subsection{Learning Strategies and Objectives}
\subsubsection{Ordinary Classification (OC)}
The methods~\cite{BRIOTA}, \cite{HoloAssist}, \cite{CaptainCook4D}, \cite{patsch2025mistsense} following this learning strategy treat both action and mistake recognition (and detection) as standard supervised classification tasks. A backbone network, such as a 3D convolutional model (e.g., I3D~\cite{carreira2017quo}) or a transformer-based architecture (e.g., TimeSformer~\cite{bertasius2021space}), is trained end-to-end using a multi-class cross-entropy loss for classifying action steps and a binary cross-entropy loss for labeling each step as correct or erroneous. Crucially, this approach requires dense, segment-level annotations, including both ground-truth action class and associated correctness label for each step. As such, it necessitates the availability of annotated sequences containing both correct and erroneous executions during training, which can be costly and time-intensive to curate, particularly in domains with complex or long-horizon tasks.

At inference, each segmented clip is classified independently, typically without access to broader task context. To mitigate prediction noise and improve temporal coherence, simple post-processing strategies such as majority voting or sliding-window smoothing are often applied across adjacent segments. Such learning objectives, inherent in structure-free methods, do not explicitly account for procedural dependencies between steps. This limits their ability to detect subtle or high-level error types, such as improper ordering of steps, repetitions, or omissions, which may only be apparent when reasoning over the entire task trajectory. Moreover, they may struggle to generalize to novel mistake types or execution variants unseen during training, given their reliance on direct supervision.

\subsubsection{One-Class Classification (OCC)} 
In this context, models for mistake recognition or detection are trained solely on samples of correct behavior, aiming to identify deviations from the learned norm during inference. This approach suits mistake and broader anomaly detection tasks, where anomalous events, such as procedural mistakes, are rare and lack explicit error examples for training. Existing works follow different strategies within this framework. Some adopt \textit{prototype learning}, clustering normal executions as reference points at test time. For instance, Lee et al.~\cite{EgoPED} learn multiple contrastive prototypes per procedural step and score incoming frames by their distance to the nearest prototype, while Seminara et al.~\cite{seminara2024differentiable} extend this idea using a differentiable task graph that models step transitions and flags off-graph deviations as errors. However, static representations struggle to capture the diversity of branching tasks. Addressing this, Huang et al.~\cite{huang2025modeling} introduce the AMNAR framework, which, similar to the graph-guided approach of Seminara et al.~\cite{seminara2024differentiable}, leverages a task graph to model valid transitions. AMNAR uses the graph to dynamically reconstruct \textit{multiple} valid action representations conditioned on the current context, ensuring that permissible variations in execution order are not misclassified as mistakes.

Other approaches explore OCC with \textit{predictive and step-aware reasoning} strategies, in which the model anticipates future procedural steps and evaluates the consistency of observed behavior against these predictions. Methods such as PREGO (in Flaborea et al.~\cite{flaborea2024prego}) and the extension TI-PREGO (in Plini et al.~\cite{plini2024ti}) utilize an OCC framework that anticipates the next step embedding based on symbolic task progression predicted by an LLM, while grounding the actual observation through a visual encoder. Discrepancies between predicted and actual embeddings are used to detect errors as they occur in real time, without ever observing mistakes during training.


\subsubsection{Task Completion (TC)}

The \textit{Task Completion} learning objective encompasses methods that evaluate model performance based on the accurate realization of an entire task, rather than relying solely on localized predictions at the frame or segment level. In this paradigm, models aim to capture the temporal dependencies and structural coherence of sub-events that collectively define a procedural sequence. The notion of a ``task" may vary in granularity: at the \textit{activity level}, it involves a sequence of high-level action steps (e.g., “crack egg”, “whisk”, “pour” to complete “make an omelet”); whereas at the \textit{action level}, it may refer to the unfolding of a single action via a trajectory of intermediate states, such as human pose configurations, hand-object interactions, or gaze fixations. Deviations between the predicted and canonical sequence, such as missing, reordered, or inconsistent steps, are interpreted as mistakes in the execution.

Training objectives in this category often include masked step prediction, sequence reconstruction, or permutation-based reasoning, encouraging the model to infer and enforce task-level consistency. Unlike conventional classification frameworks, TC approaches do not require explicit error labels; correctness is inferred from whether the predicted trajectory conforms to a learned procedural schema. For instance, Narasimhan et al.~\cite{narasimhan2023learning} learn structured task representations by predicting randomly masked steps in instructional videos, allowing the model to identify steps that break the expected temporal or logical flow.
In a complementary formulation, a subset of methods approach this objective via anticipation-based reasoning, where correctness is evaluated by comparing the model’s recognized step with the expected next step, as inferred from a structured procedure. PREGO~\cite{flaborea2024prego} and TI-PREGO~\cite{plini2024ti} formulate mistake detection as a joint problem of \textit{step recognition} and \textit{step anticipation}, where a mismatch between anticipated and recognized steps signals a deviation from the canonical trajectory of the action-steps of the activity. 

\begin{table*}[t]
\centering
\resizebox{\textwidth}{!}{
\begin{tabular}{l|c|c|l|c|l|c|c|c|c|c|c|c|l}
\hline
\textbf{Method} & \textbf{Task} & \textbf{Error} & \textbf{Error} & \textbf{Error}&      \textbf{Dataset}          & \textbf{View} & \textbf{Learning} & \textbf{Modality/Input} & \textbf{Superv.} & \textbf{Procedural} & \textbf{Online} & \textbf{Video} &\textbf{ Action \& Activity} \\
 &              & \textbf{Type} & \textbf{Granularity} & \textbf{Metric}&  &  & \textbf{Strategy} &        &  & \textbf{Structure} &  & \textbf{Trim.} &\textbf{Learning Method} \\
\toprule
\hline
Moriwaki et al.~\cite{BRIOTA} & MD & PEs, & Coarse & TALM,  & BRIO-TA~\cite{BRIOTA} & Ego &  OC & RGB & Fully & STM & \ding{55} & \checkmark & I3D~\cite{carreira2017quo} + MS-TCN~\cite{farha2019ms}\\
\textit{ICIP'22} & & EEs & & ASA&  &  & & & & &  & & \\ 
\hline 
Ding et al.~\cite{ding2023mistakecountsassembly}  & MD & PEs & Coarse,  & CPM & Assembly101~\cite{Assembly101} & NV &  TC & Text (AL) & Full & STM, SKG & \ding{55} & \checkmark & TempAgg~\cite{sener2020temporal} \\
\textit{arXiv'23} & & & Fine & &  &  & & & & &  & & \\ 
\hline 
Narasimhan et al.~\cite{narasimhan2023learning} & MD & PEs& Coarse& CPM& COIN~\cite{COIN} & Ego/ & TC & RGB & Weak & TDM & \ding{55} & \checkmark & VideoTaskformer~\cite{narasimhan2023learning} \\
\textit{arXiv'23} & & & & &  & Exo & & & & &  &  & \\ 
\hline 
Wang et al.~\cite{HoloAssist}  & MD & PEs, & Coarse&CPM& HoloAssist~\cite{HoloAssist}& Ego & OC & RGB,  & Full & STM & \ding{55} & \checkmark & TimeSformer~\cite{bertasius2021space} \\
\textit{ICCV'23}  & EMP & EEs & & &  &  & & HP, EG & & &  & & \\ \hline
Ghoddoosian et al.~\cite{ATA} & MD & PEs& Coarse & CPM, & ATA~\cite{ATA}, CSV~\cite{CSV} & Exo & OC, OCC,  & RGB & Weak & TDM & \ding{55} & \checkmark & I3D~\cite{carreira2017quo} + OadTR~\cite{wang2021oadtr} \\
\textit{ICCV'23} & & & &TALM &  &  & VTA & & & &  &  & + Viterbi variant~\cite{ATA}\\  
\hline
Schoonbeek et al.~\cite{IndustReal} & MR & PEs& Coarse& POS& IndustReal~\cite{IndustReal} & Ego & TC, OCC & RGB, VL,  & Weak & TDM & \checkmark & Both & MViTv2~\cite{li2022mvitv2}  \\
\textit{WACV'24} & & & & CPM (F1)&  &  & & Stereo & & &  &  &\\ 
\hline 
Peddi et al.~\cite{CaptainCook4D} & MTR, & PEs, & Coarse, & CPM& CaptainCook4D~\cite{CaptainCook4D} & Ego & CC & RGB & Full & SFM (\textit{MTR}, & \ding{55} & \checkmark & 
Omnivore~\cite{girdhar2022omnivore} \\ 
\textit{NeurIPS'24} &  EMP, &EEs & Fine & &  &  & & &  & \textit{EMP}), STM (\textit{D}) &  &  &  + ActionFormer~\cite{zhang2022actionformer}\\ 
 & MD  & &  & &  &  & & & & &  &  &\\ \hline
Lee et al.~\cite{EgoPED} & MD&PEs&Coarse&EDA, CPM (AUC), & EgoPER~\cite{EgoPED},  ATA~\cite{ATA} & Ego & OCC & RGB & Full& SFM & \ding{55} & \ding{55} & (I3D\cite{carreira2017quo}+FasterRCNN~\cite{ren2016faster}  \\
\textit{CVPR'24} & & & &O-IoU, TALM & HoloAssist~\cite{HoloAssist} &  & & & & &  &  & +GCN) + ActionFormer~\cite{zhang2022actionformer} \\ 
\hline 
Flaborea et al.~\cite{flaborea2024prego} & MD& PEs&  Coarse&CPM & Assembly101-O~\cite{flaborea2024prego},  & Ego & TC, OCC & RGB & Weak& TDM & \checkmark & Both & OadTR~\cite{wang2021oadtr} + Llama 2.0 \\
\textit{CVPR'24} & & & &  & Epic-Tent-O~\cite{flaborea2024prego} &  & & & & &  & & \\ 
\hline 
Seminara et al.~\cite{seminara2024differentiable}  & MD& PEs& Coarse& CPM & Assembly101-O~\cite{flaborea2024prego},  & Ego & TC, OCC & RGB & Weak & PTG & \checkmark & Both & MinoRoad~\cite{an2023miniroad} +\\ 
\textit{NeurIPS'24} & & & & &Epic-Tent-O~\cite{flaborea2024prego} + Task  &  & & & & &  & & Graph Transformer~\cite{seminara2024differentiable}\\ 
\hline 
Haneji et al.~\cite{EgoOops} & MD& PEs& Coarse& TALM & EgoOops~\cite{EgoOops} & Ego & VTA, TC & RGB & Fully & TDM, STM & \ding{55} & \checkmark & StepFormer~\cite{dvornik2023stepformer} \\
\textit{arXiv'24} & & & & &  &  & & & & &  & & + Drop-DTW~\cite{dvornik2021drop}\\ 
\hline 
Plini et al.~\cite{plini2024ti}  & MD& PEs&Coarse& CPM & Assembly101-O~\cite{flaborea2024prego},  & Ego & TC, OCC & RGB & Weak & TDM & \checkmark & Both& MiniRoad~\cite{an2023miniroad}  \\
\textit{arXiv'24} & & & & & Epic-Tent-O~\cite{flaborea2024prego} &  & & & & &  & & + Llama 3.1 (CoT)\\ 
\hline 
Mazzamuto et al.~\cite{mazzamuto2025gazing}  & MD& EEs & Coarse& CPM & HoloAssist~\cite{HoloAssist},  & Ego & OCC & RGB, & \ding{55} &  SFM& \checkmark & Both & Gaze-centered Transformer \\
\textit{CVPR'25} & & &  & & Epic-Tent~\cite{EPICTENT} &  & &  EG& & &  & & Autoenc. + MoCoDAD~\cite{flaborea2023multimodal}\\ 
\hline 
Huang et al.~\cite{huang2025modeling}  & MD& PEs & Coarse & EDA, & EgoPER~\cite{EgoPED}, HoloAssist~\cite{HoloAssist},  & Ego & OCC & RGB, & Weak &  PTG& \ding{55} & Both & I3D~\cite{carreira2017quo} + ActionFormer~\cite{zhang2022actionformer} \\
\textit{CVPR'25} & & & & CPM (Prec, AUC)& Captain-Cook4D~\cite{CaptainCook4D} &  & & TG& & &  & & +Graph Structure Operations\\ 
\hline
Hou et al.~\cite{ICME}  & MD& EEs& Coarse& EDA & HoloAssist~\cite{HoloAssist},  & Ego & OCC & RGB & Weak & STM& \ding{55} & Both & (ActionFormer~\cite{zhang2022actionformer}+ CDC)+ \\
\textit{ICME'25} & &PEs**& & CPM (AUC)& EgoPER~\cite{EgoPED} &  & & & & &  & & GMM+Gaussian Smoothing Module\\ 
\hline 
Kung et al.~\cite{kung2025changed} & MD& PEs, & Coarse &EDA, & EgoPER~\cite{EgoPED} & Ego & VTA, TC & RGB, & Weak & PTG & \ding{55} & \ding{55}& Llama 3 +  ASFormer~\cite{yi2021asformer} \\
\textit{arXiv'25} & & EEs & &CPM (AUC) &  &  & & Text & & &  & & + HierVL~\cite{ashutosh2023hiervl}\\ 
\hline
Lee and Elhamifar~\cite{lee2025error} & MD& PEs &  Coarse& TALM, & EgoPER~\cite{EgoPED}, & Ego & TC& RGB, & Weak & PTG & \ding{55} & Both& DiffAct~\cite{liu2023diffusion} + G2Vid~\cite{dvornik2022flow}\\
\textit{ICCV'25} & & EEs &Fine & CPM& CaptainCook4D*\cite{CaptainCook4D}  &  & & TG& & &  & & + (VideoCLIP~\cite{Xu2021VideoCLIPCP} + GPT4o mini)\\ 
\hline
  & &  &  &  &   &  & & &  &  &  &  & (ViT~\cite{dosovitskiy2020image} + Q-Former~\cite{li2023blip}) +\\
Patsch et al.~\cite{patsch2025mistsense} & MD-e & PEs, & Coarse,& CPM (F1) & Epic-Tent-O~\cite{flaborea2024prego}, & Ego & OC,&  RGB,& Fully&STM & \checkmark  &Both & (HardFormer~\cite{shamil2024utility} + Q-Former~\cite{li2023blip}) \\ 
\textit{ICCV'25} & & EEs & Fine& NGM& HoloAssist~\cite{HoloAssist}&  & VTA & HP& & &  & & + Llama 2.0\\ 
\hline
Li et al.~\cite{li2025mistake} & (ST) MD&  EEs& Coarse & CPM, TALM, & EK-M~\cite{li2025mistake}, & Ego & VTA& RGB & Fully & STM & \ding{55}& \checkmark& SRL~\cite{gardner2018allennlp} + InternVideo2~\cite{wang2024internvideo2}\\
\textit{arXiv'25} & & PEs** &Fine & OBJM& Ego4D-M~\cite{li2025mistake}  &  & & Text& & &  & & + Transformer Heads \\ 
\hline

\end{tabular}}
\vspace*{0.10cm}
\caption{\textbf{Method taxonomy across key design factors.} \underline{Tasks:} \{MR–mistake recognition, MTR–mistake type recognition, EMP–early mistake recognition, MD–mistake detection (temporal), (ST)MD - spatio-temporal MD, -e: supports explainability\}. \underline{Modalities:} \{AL–action labels, HP–hand pose, EG–eye gaze, VL–visible light, TG-task graph\}. \underline{Procedural structure:} \{STM–stepwise model, TDM–template-driven model, SKG–semantic knowledge graph, PTG–procedural task graph, SF–structure-free\}. \underline{Error metrics:} \{CPM–classification metrics (Acc, Prec, Rec, F1, AUC), TALM–temporal localization metrics (Edit Distance, IoU, F1$@0.5$), EDA–error detection accuracy, POS–procedure order similarity, ASA–anomaly section accuracy, OIoU–omission IoU, NGM: n-gram metrics (BLEU, ROUGE-L, CIDEr).\}. \underline{Misc.:} \{NV: non-vision method, OBJM-object detection metrics, GCN–graph convolutional networks,  ACoT–Automatic Chain of Thought~\protect\cite{zhang2022automatic}, TAS-Temporal Action Segmentation, CDC - Causal Dilated Convolution, GMM- Gaussian Mixture Model\}. \underline{Note that:} \textit{For works introducing novel datasets, the top-performing baseline is listed in the last column. Overall, for every work we list top-performing model configuration.} *: evaluated on a subset of the dataset.**:\protect\cite{ICME} does not explicitly model procedural structure; such errors are detected when manifested as deviations in the per-step visual distribution.}
\label{tab:method-taxonomy}
\end{table*}

\subsubsection{Video-Text Alignment as a Supervisory Signal (VTA)} 

VTA-based approaches utilize procedural text (e.g., step descriptions from manuals, task guides, or LLM-generated scripts) as a supervisory signal to guide action detection, anticipation, and activity understanding~\cite{kahatapitiya2024victr, Chen_2024_CVPR, li2024topa}. These methods typically segment untrimmed videos into candidate action clips by aligning them to textual templates using techniques such as dynamic time warping, cross-modal attention, or contrastive learning between video and language embeddings, then pass the aligned segments to classification or matching modules to predict the activity or next action.

In mistake analysis, VTA enables assessing step correctness by evaluating alignment strength or semantic consistency between predicted visual content and the textual description of correct behavior. Its adoption remains limited due to scarce datasets like EgoOops~\cite{EgoOops} and CaptainCook4D~\cite{CaptainCook4D}, which provide paired instructional text and annotated mistakes. In EgoOops, Haneji et al.~\cite{EgoOops} align egocentric video segments with procedural transcripts. StepFormer++ predicts the current step and localizes mistakes by detecting misalignments between visual input and expected procedural steps, using cross-attention to learn temporal and semantic alignment between video tokens and step embeddings. To address data limitations, Kung et al.~\cite{kung2025changed} propose a counterfactual VTA framework for mistake analysis using Ego4D~\cite{grauman2022ego4d}. While Ego4D lacks error annotations, it provides rich activity- and action-level captions. LLaMA 3.2 8B is used to generate counterfactual descriptions at action and activity levels, describing missing, misordered, or incorrectly executed steps. These synthetic error-augmented captions are paired with correct video clips for contrastive training of HierVL~\cite{ashutosh2023hiervl}, a hierarchical video-language model.

Moving beyond sentence-level alignment, Li et al.~\cite{li2025mistake} in \textit{MisFormer}, which utilizes Semantic Role Labeling (SRL) to decompose procedural instructions into fine-grained components (e.g., Predicate (verb), Object). Unlike previous VTA approaches that treat the text as a single embedding, MisFormer employs a dual-branch transformer that cross-attends these specific role tokens against video patches. This enables the model to determine specifically \textit{which} part of the instruction was violated (e.g., correct action (verb) but wrong object). Complementing these attribution-focused methods, Patsch et al.~\cite{patsch2025mistsense} leverage VTA for post-hoc explanation rather than just detection. Their MistSense framework employs a gating mechanism that, upon detecting an error, projects video-based features into the embedding space of a Large Language Model (LLM). This allows the system to generate natural language descriptions of the error (e.g., ``\textit{The person is mounting a screw in a wrong way}''), bridging the gap between binary detection and interpretable user feedback.

\subsection{Supervision Levels in Mistake Analysis}

\subsubsection{Fully Supervised Approaches}

Fully supervised approaches rely on datasets with explicit error annotations at a fine-grained level. In this setting, mistakes are annotated with precise spatial and temporal localization, typically at the frame or segment level, enabling models to learn the error type and also its exact temporal occurrence in the video. Most methods following this paradigm operate in an offline setting~\cite{CaptainCook4D}, \cite{Assembly101}, \cite{HoloAssist}, where detailed annotations are used to supervise both step segmentation and mistake classification.  Ding et al.~\cite{ding2023mistakecountsassembly} leverage fine-grained labels to construct spatiotemporal knowledge graphs guiding mistake reasoning. Similarly, Mirowaki et al.~\cite{BRIOTA}, in the BRIO-TA dataset, provide segment-level annotations indicating correct or erroneous execution, allowing models to be trained using standard supervised classification objectives at the action segment level.

\subsubsection{Weakly-Supervised Methods}
Weak supervision is a learning strategy employed to reduce annotation effort, characterized by two complementary perspectives: (1) the methodological design leveraging auxiliary supervision signals and (2) the granularity of annotations available during training.

In mistake analysis, the first perspective concerns the absence of explicit mistake annotations, compensated by exploiting the procedural structure (actions and their transitions) of the activity. This is exemplified by methods leveraging global step order to detect execution inconsistencies~\cite{grauman2024ego, ATA, IndustReal, narasimhan2023learning, lee2025error}. Although these methods do not require annotated mistake labels for training, they still rely on step-level annotations, i.e. knowledge of actions and their order. The second perspective relates to annotation granularity, where weak supervision labels mistakes only at the video level (e.g., indicating a mistake’s presence) without specifying when it occurs. In such cases, models must implicitly localize deviations based on high-level error cues (e.g., “action skipped”).

Grauman et al. in Ego-Exo4D~\cite{grauman2024ego} explore a weak mistake detection strategy with two variants depending on supervision granularity: (1) \textit{instance-level supervision}, where both video segments and key-step labels are provided during training and inference, analogous to action recognition, and (2) \textit{procedure-level supervision}, where training and inference use unlabeled segments along with a taxonomy of procedure-specific key-step names. In the second approach, key-step assignments are generated via a pre-trained video-language model, providing pseudo-labels and exemplifying self-supervised weak supervision. Similarly, Ghoddoosian et al.~\cite{ATA} propose a weakly-supervised method for instructional mistake detection using only activity transcripts without frame-level labels or error examples. Supervision is limited to expected action sequences, with no temporal error information. Their method infers frame-wise action boundaries by aligning predictions to the transcript using a loss that enforces consistency with valid sequences while maximizing separation from invalid ones. Narasimhan et al.~\cite{narasimhan2023learning}, in \textit{VideoTaskformer}, operate under weak supervision differently: instead of aligning to a transcript, the model learns procedural knowledge by predicting masked steps within instructional sequences. Trained with step-level annotations (e.g., “crack egg”, “whisk”, “pour”), without step timing or correctness information, it infers missing steps based on surrounding context, acquiring a representation of procedural structure that identifies absent or out-of-order steps. In more recent works, Lee and Elhamifar~\cite{lee2025error} andJ Huang et al.~\cite{huang2025modeling} utilize  structure-driven weakly supervised methods that leverage a predefined task graph as indirect supervision. Unlike transcript-alignment or masked-step prediction approaches, these methods do not require frame-level step labels or error annotations; instead, they perform joint step alignment and error recognition by optimizing a dynamic programming objective constrained by the graph topology. This framework allows both frame~\cite{lee2025error} and node~\cite{lee2025error,huang2025modeling} skipping, enabling the model to infer deviations from canonical execution without relying on densely annotated training data. 

A complementary form of weak supervision is introduced by the PECC framework~\cite{ICME}, which requires only step-level annotations of correct executions and no mistake labels. Unlike transcript-based alignment methods~\cite{ATA}, masked-step prediction approaches~\cite{narasimhan2023learning}, or graph-constrained frameworks~\cite{lee2025error,huang2025modeling}, PECC does not rely on external procedural structure, textual cues, or predefined transition rules. Instead, it learns per-step feature distributions by training a temporal action segmentation backbone on correct sequences and fitting Gaussian Mixture Models in a one-class manner. Weak supervision here is defined purely through exposure to correct behavior, without auxiliary signals such as transcripts, task graphs, or LLM-generated pseudo-labels. PECC treats each step (action) independently, detecting errors as low-likelihood deviations in the learned feature space. This makes the approach effective for executional anomalies but limits its ability to capture higher-level procedural mistakes, such as step omissions or misordering, that other weakly supervised methods address through explicit global structure or sequence-level reasoning.

Finally, a recent sub-line of weakly-supervised approaches~\cite{flaborea2024prego, plini2024ti} departs from previous formulations and defines mistake detection as a combination of \textit{step recognition} and \textit{step anticipation}. These methods exploit the structured nature of task transcripts as supervision signals, without requiring any frame-level action or mistake annotations. Notably, they leverage pre-trained LLMs as reasoning agents to infer expected next steps based on procedural understanding.

\subsubsection{Unsupervised Methods} 

Unsupervised methods for mistake recognition or detection aim to identify deviations in action execution without labeled mistakes, annotated boundaries, or knowledge of an action’s procedural role. They assume only correct executions are available during training, with mistakes manifesting as deviations, statistically or structurally distinct from valid behavior. Accordingly, such methods learn compact representations or generative models of correct task performance, flagging as outliers those instances that cannot be well explained, reconstructed, or predicted by the model.

Existing unsupervised methods follow strategies centered on \textit{predictive modeling}, where states of modalities (e.g., gaze, human 2D/3D pose) are predicted and mistakes inferred from large prediction deviations. They rely on latent modeling of correct trajectories, often as feature embeddings, where deviations from learned distributions are flagged as mistakes/anomalies. \textcolor{black}{This category of mistake recognition/detection methods is related to VAD methods (see Section~\ref{Mis_VAD}). However, even though they do not explicitly model task semantics, action steps, or procedural knowledge, unsupervised mistake analysis methods are applied to procedural activities (e.g., cooking, object assembly), where mistakes are subtle and often arise from missing, improperly ordered, or poorly executed steps. In contrast, VAD methods with similar approaches~\cite{sato2023prompt, markovitz2020graph, noghre2024exploratory, stergiou2024holistic} focus on coarse distinct actions with no procedural context, usually in surveillance applications}. Mazzamuto et al.~\cite{mazzamuto2025gazing} leverages egocentric gaze prediction as an unsupervised proxy for task regularity. It defines a novel gaze completion task and measures deviations between predicted and observed gaze to detect mistakes online, without using any labels or manual annotations. To capture this, they formulate the gaze completion task with a transformer-based model that learns to predict future gaze trajectories based on past egocentric video and gaze history. During inference, discrepancies between the predicted gaze and the actual observed gaze are interpreted as anomalies, with large deviations signaling possible mistakes or disruptions in task execution.

\section{Challenges and Future Directions}

Despite significant progress in procedural activity understanding, a number of challenges and open problems remain. Addressing them is critical for advancing mistake analysis, particularly in developing methods that are robust, interpretable and generalizable. We outline the major open problems in the field and highlight promising research directions.

\subsection{Challenges in Mistake Analysis}
\label{chal_probs}
\noindent \textbf{Datasets, modality \& evaluation bottlenecks:} Despite growing interest in mistake analysis for procedural activities, the limited number and scale of comprehensive datasets remain major bottlenecks. Most existing video datasets target action or activity recognition, detection, or anticipation, typically assuming correct task execution. As a result, mistake annotations 
(when available) are often coarse, restricted to binary video-level labels (mistake vs. correct), and omit key aspects such as temporal boundaries, causal origins, or semantic types (e.g., omission, misordering). The scarcity of temporally localized and semantically disambiguated annotations hampers model development and evaluation, while annotation protocols remain labor-intensive and domain-specific, particularly for subtle or ambiguous errors. Beyond annotation gaps, datasets exhibit limited modality coverage and biased recording perspectives. Modalities such as gaze, object state, 2D/3D pose, or synchronized audio can enhance error analysis but are often absent due to sensor or collection constraints. Most datasets adopt egocentric viewpoints, especially for household or instructional tasks, limiting applicability to industrial contexts where exocentric or hybrid ego–exo views are more suitable. The lack of standardized evaluation metrics further complicates comparison: most works rely on general-purpose classification or segmentation metrics (e.g., Acc., F1-score, IoU) that poorly capture the challenges of mistake analysis tasks, hindering cross-method benchmarking. 
\vspace{0.1cm}

\noindent \textbf{Ambiguity over permissible variations \& actual errors:} A key problem in procedural mistake analysis is distinguishing legitimate execution variations from true errors. Human procedures are inherently variable, often allowing multiple valid action sequences, interaction styles, or tool-use alternatives achieving the same goal without compromising quality. Annotation challenges stem from limited consensus on what constitutes a mistake versus a permissible variation, driving high labeling costs and inter-annotator disagreement, particularly when procedural flexibility is high or correctness criteria implicit. Consequently, existing datasets often omit such nuance or adopt simplified taxonomies overlooking real-world task complexity. Modeling approaches typically rely on distributional regularities learned from data, lacking explicit representations of task goals, procedural constraints, or causal dependencies. This leads to misclassification of rare but correct behaviors and missed detections of context-dependent violations requiring deeper task reasoning. Addressing this necessitates models to reason about action affordances, object roles, and the causal effects of omissions or order errors. Recent video-language models~\cite{tang2025video} jointly encode temporal and semantic context and can be prompted with structured task goals or contextual cues. Yet precise error attribution remains difficult, especially for finer distinctions and task-specific correctness rules. Vision–language approaches like PREGO~\cite{flaborea2024prego} and TI-PREGO~\cite{plini2024ti} show promise in identifying procedural errors, though executional ones remain underexplored. Their sensitivity to prompt phrasing and contextual formulation~\cite{zhou2022learning} further raises robustness concerns in real-world settings.

\vspace{0.1cm}
\noindent \textbf{Disjoint embedding spaces between LLMs \& video encoders:} 
A key limitation in current mistake analysis frameworks lies in the lack of task-level reasoning, where models fail to encode procedural goals, causal dependencies, and object affordances, resulting in systems that primarily detect low-level visual deviations without understanding whether an observed deviation truly constitutes a mistake~\cite{mazzamuto2025gazing}. Video-Language Models (Video LLMs)~\cite{flaborea2024prego,plini2024ti} have emerged as a promising direction, integrating the semantic richness of LLMs with the temporal and spatial modeling capacity of video encoders. However, representational misalignment between the two modalities (vision, language) remains a core challenge. Current designs rely on dual-stream architectures with shallow fusion, handcrafted prompts, or minimal supervision of cross-modal correspondences. As a result, they struggle to ground complex procedural logic or contextual correctness in visual data~\cite{rawal2023dissecting} and remain limited to detecting procedural mistakes, which require alignment with predefined step sequences, rather than executional mistakes, which require fine-grained perception of action quality and context-dependent deviations.


\subsection{Future Directions in Mistake Analysis}
\noindent\textbf{Lifelong learning \& open-vocabulary recognition:} A major limitation of current methods is the reliance on closed-world training setups, where action-object classes, and task flows are assumed fixed and fully known during training. However, real-world procedures are dynamic, evolving across domains, involving diverse object-tool configurations, and introducing novel behaviors or error patterns unseen during training. This motivates a shift toward lifelong learning and open-vocabulary recognition frameworks for more robust, scalable, and generalizable analysis. Lifelong learning paradigms, particularly for video understanding (e.g., \cite{gowda2024continual}), enable models to incrementally learn from streaming video while retaining performance on prior tasks. Such adaptability is crucial for vision-based mistake analysis, especially for executional mistakes, where new action styles may arise due to tool substitutions. Complementing continual learning, open-vocabulary recognition allows models to describe and detect mistakes beyond closed taxonomies, enabling zero-shot recognition of novel mistake classes. Future methods could draw inspiration from action understanding works that generalize beyond fixed label sets using vision-language alignment and compositional prompts in multi-modal large language models (e.g., \cite{gupta2024open, chatterjee2023opening}).
\vspace*{0.03cm}

\noindent\textbf{Explainable analysis via symbolic reasoning integration with video models:} A key challenge in vision-based mistake analysis is the internal mechanisms that guide model decisions, especially when detecting subtle deviations in procedural tasks. Current end-to-end neural models often operate as black boxes, offering little insight into why a given action is classified as a mistake. Such explainability should be considered a foundational principle and a critical requirement due to the inherent flexibility of human-performed tasks. Many procedures allow for multiple valid execution paths, making it essential for models to differentiate between permissible variations and true violations of task constraints. A promising direction towards this goal is integrating symbolic reasoning, such as procedural constraints, causal rules and temporal logic, with neural video perception. Mistake analysis could benefit from neuro-symbolic designs that (a) parse observed video into symbolic representations of steps (a concept briefly explored in PREGO~\cite{flaborea2024prego} and TI-PREGO~\cite{plini2024ti} for aligning observed actions with expected procedural steps) and object states~\cite{gouidis2024enabling}, (b) enforce procedural logic via symbolic reasoners or temporal logic automata~\cite{choi2024towards}, and (c) ground discrepancies (both procedural and executional) as violations of explicit task rules or specific semantic roles (e.g., predicate or object mismatches~\cite{li2025mistake}). Such hybrid systems would enhance detection accuracy and offer interpretable explanations as well as zero-shot generalization across new procedures. Regarding explainability, \cite{patsch2025mistsense} demonstrate the potential of using LLMs to generate intuitive natural language error descriptions directly from visual features; however, ensuring these generative explanations remain factually grounded and structurally verifiable, remains a critical challenge that necessitates the integration of explicit symbolic constraints. Furthermore, the reliance on standard captioning metrics (e.g., BLEU, CIDEr)  to evaluate these systems~\cite{patsch2025mistsense} is arguably inadequate for mistake analysis; such metrics quantify textual overlap rather than the causal correctness or interventional utility of the diagnosis, underscoring the need for new, logic-grounded evaluation protocols.
\vspace*{0.01cm}

\noindent\textbf{Understanding error dependencies \& propagation:} Building on interpretable representations for procedural understanding, the next step is to move beyond  detection and examine how errors influence downstream actions within task sequences. In real-world industrial or medical settings, errors rarely occur in isolation; initial deviations can propagate, altering execution and causing more severe failures. Understanding these inter-error dependencies is essential for systems that not only detect mistakes but anticipate and mitigate cascading effects. While early works like AMNAR~\cite{huang2025modeling} use graph structures to predict valid next-steps based on current history, future methods must extend this to explicitly learn the likelihood of specific error transitions (e.g., Error A increases the probability of Error C) to enable early interventions preventing cascading failures. Integrating predictive capabilities into real-time mistake detection could enhance automated quality control and prevention. Recent synthetic-domain works like~\cite{hong2021transformation}, can model explicit visual state transitions, providing a foundation for inferring causal chains of mistakes via object state transitions in real-world procedural tasks.
\vspace*{0.03cm}

\noindent\textbf{Enhancing datasets with counterfactual state transitions and synthetic mistakes:} Procedure-aware video understanding and mistake recognition require modeling both actual and hypothetical scene state transitions, capturing action-induced transformations and plausible deviations from incomplete or erroneous actions. Several works have highlighted the link between action or activity recognition and object/scene state changes~\cite{ChopANDLearn, wang2016actions, Bacharidis_2023_ICCV, souvcek2022look, niu2024schema}. A promising direction for advancing mistake analysis involves augmenting procedural datasets with synthetic state-change sequences and counterfactual exemplars~\cite{kung2025changed, GenHowTo, chatzi2024counterfactual, souvcek2025showhowto}, enabling models to reason about the causal and semantic impact of actions by grounding visual representations in expected and unexpected outcomes. This fosters richer temporal representations that distinguish correct progressions from deviations or failures, even in subtle or ambiguous cases. Recent works by Lee and Elhamifar~\cite{lee2025error}, as well as Kung et al.~\cite{kung2025changed} showed that, given prior knowledge of an activity’s execution protocol, one can use structured procedural knowledge bases (e.g., WikiHow~\cite{koupaee2018wikihow}) or LLM-generated textual descriptions to define expected pre- and post-action states and hypothetical post-conditions for potential failures. These textual representations can then be aligned with visual features to reinforce plausible state transitions while decoupling them from counterfactual ones via contrastive learning, thereby strengthening the visual grounding of procedural semantics. Complementing these generative approaches, Li et al.~\cite{li2025mistake} demonstrated a retrieval-based data augmentation strategy which allows the automatic construction of semantically attributed mistake video samples by systematically cross-matching instruction texts with video segments of disparate actions from existing large-scale corpora (e.g., pairing a ``pick up hammer'' instruction with a ``pick up bolt'' video) allowing for retrieval-based clip switching. However, such strategy fails to maintain visual consistency with the preceding action clips of the activity sample, as well as preserving scene semantics. This discontinuity disrupts the temporal coherence required to model realistic error propagation and causal dependencies in long-horizon activities.

\section{Conclusions}
Vision-based mistake analysis in procedural activities is a rapidly evolving field with potential to improve task performance, safety, and efficiency across domains. This paper reviews existing methods, datasets, and evaluation metrics, outlining challenges and opportunities in detecting and predicting procedural and executional errors. By categorizing approaches by procedural structure, supervision, and learning strategies, we offer a structured view of the state of the art. Key challenges remain, such as the ambiguity between permissible variations and true mistakes, limited comprehensive datasets, and the need for models that reason about error dependencies and propagation. Addressing these requires innovations such as neuro-symbolic reasoning, leveraging large language models for procedural understanding, and enriching datasets with counterfactual transitions and synthetic mistakes. Future research should emphasize explainable mistake analysis, stronger cross-task generalization, and real-time predictive detection, unlocking new possibilities for human–robot collaboration, automation, and assistive technologies, and ultimately transforming how mistakes are understood and mitigated in structured activities.

\section*{Acknowledgments}
This work was conducted within the framework of the Action ‘Flagship Research Projects in challenging interdisciplinary sectors with practical applications in Greek industry,’ implemented via the National Recovery and Resilience Plan Greece 2.0, funded by the European Union – NextGenerationEU (project: TAEDR-0535864). It was also co-funded by the European Union (EU - HE Magician – Grant Agreement 101120731) and the VMware University Research Fund.

\textcolor{black}{We acknowledge the use of ChatGPT (GPT-5) and Gemini (2.5 Flash) for creating components of the figures and assisting with language corrections during manuscript editing.}

\bibliographystyle{IEEEtran}   
\bibliography{mybibliography}   
\vfill

\end{document}